\documentclass[]{interact}

\usepackage{epstopdf}% To incorporate .eps illustrations using PDFLaTeX, etc.
\usepackage{graphicx}
\usepackage{bm}
\usepackage{amssymb}
\usepackage{multirow}
\usepackage{amsmath}
\usepackage{caption}
\usepackage{tablefootnote}
\usepackage{xcolor}
\usepackage{subfigure}

\usepackage[numbers,sort&compress]{natbib}% Citation support using natbib.sty
\bibpunct[, ]{[}{]}{,}{n}{,}{,}% Citation support using natbib.sty
\renewcommand\bibfont{\fontsize{10}{12}\selectfont}% Bibliography support using natbib.sty
\makeatletter% @ becomes a letter
\def\NAT@def@citea{\def\@citea{\NAT@separator}}% Suppress spaces between citations using natbib.sty
\makeatother% @ becomes a symbol again

\theoremstyle{plain}% Theorem-like structures provided by amsthm.sty

\theoremstyle{definition}

\theoremstyle{remark}

\begin{document}

\title{Measuring the Similarity between Materials with an Emphasis on the Materials' Distinctiveness}

%\author{
%\name{A.~N. Author\textsuperscript{a}\thanks{CONTACT A.~N. Author. Email: latex.helpdesk@tandf.co.uk} and John Smith\textsuperscript{b}}
%\affil{\textsuperscript{a}Taylor \& Francis, 4 Park Square, Milton Park, Abingdon, UK; \textsuperscript{b}Institut f\"{u}r Informatik, Albert-Ludwigs-Universit%\"{a}t, Freiburg, Germany}
%}

\author{
\name{
Tran-Thai Dang\textsuperscript{a},
Tien-Lam Pham\textsuperscript{a,b,c},
Hiori Kino\textsuperscript{b,c},
Takashi Miyake\textsuperscript{b,c,d}, and
Hieu-Chi Dam\textsuperscript{a,b,e}\thanks{CONTACT Hieu-Chi Dam. Email: dam@jaist.ac.jp}
}
\affil{
\textsuperscript{a}Japan Advanced Institute of Science and Technology,1-1 Asahidai, Nomi, Ishikawa 923-1211, Japan;
\textsuperscript{b}Center for Materials Research by Information Integration, National Institute for Materials Science, 1-2-1 Sengen, Tsukuba, Ibaraki 305-0047, Japan;
\textsuperscript{c}ESICMM, National Institute for Materials Science, 1-2-1 Sengen, Tsukuba, Ibaraki 305-0047, Japan;
\textsuperscript{d}CD-FMat, AIST, 1-1-1 Umezono, Tsukuba 305-8568, Japan;
\textsuperscript{e}JST, PRESTO,\\4-1-8 Honcho, Kawaguchi, Saitama 332-0012, Japan
}
}

\maketitle

\begin{abstract}
In this study, we establish a basis for selecting similarity measures when applying machine learning techniques to solve materials science problems. This selection is considered with an emphasis on the distinctiveness between materials that reflect their nature well. We perform a case study with a dataset of rare-earth transition metal crystalline compounds represented using the Orbital Field Matrix descriptor and the Coulomb Matrix descriptor.  We perform predictions of the formation energies using k-nearest neighbors regression, ridge regression, and kernel ridge regression. Through detailed analyses of the yield prediction accuracy, we examine the relationship between the characteristics of the material representation and similarity measures, and the complexity of the energy function they can capture. Empirical experiments and theoretical analysis reveal that similarity measures and kernels that minimize the loss of materials' distinctiveness improve the prediction performance.
\end{abstract}


\section{Introduction}\label{sec:intro}
A small change in the chemical composition or structure of materials can lead to a significant change in the properties of materials. For example, differences in the chirality of a honeycomb network of carbon atoms can lead to a distinctive difference in physical properties of nanotubes. The distinctiveness of materials, which results in the diversity of materials in the nature, is a main characteristic of the material data. Thus, this feature needs to be represented in a metric that allows for a comparison of materials in a reliable, efficient, and useful way.

The main target of machine learning systems when mining material data is to determine a likely function $f(x)$, which indicates the relation between the materials' attributes and their physical properties. Typically, these systems include two main components: (i) data representations (i.e., descriptors); and (ii) operators (including similarity measures between materials and learning methods) for mining the physical and chemical properties of materials. These components are designed with the aim of reflecting domain knowledge and the nature of material data.

To render computational methods tractable for materials in datasets, the geometrical, topological, or electronic characteristics of the materials need to be represented in form of numerical variables. Descriptors commonly encode the information of a material $A$ by a vector $\vec{x}_A = (x_A^1, x_A^2,..., x_A^m)$ whose number of dimensions, $m$, and values in each dimension depend on the information selected to describe the materials with a specific purpose for mining tasks. To represent material structures, several descriptors have been proposed.
Behler \textit{et al.} utilized atom-distribution-based symmetry functions to represent the local chemical environment of atoms~\cite{behler2011atom}. Rupp \textit{et al.} proposed the Coulomb matrix (CM), which represents materials via the Coulomb repulsion between all possible nuclei in the material~\cite{rupp2012fast}. In addition, Isayev \textit{et al.} used the band structure and density of states (DOS) fingerprint vectors as descriptors of materials to visualize material space~\cite{isayev2015materials}. Zhu \textit{et al.} introduced another fingerprint representation for crystals and used this to define the configurational distance between crystalline structures~\cite{zhu2016fingerprint}. Pham \textit{et al.} proposed a descriptor for encoding atomic orbital information, called the orbital field matrix (OFM)~\cite{lam2017machine,pham2018learning}.

Similarity measures are mathematically implemented as scalar valued functions that take two vectors representing materials $A$ and $B$ as input: $S(A, B) = S(\vec{x}_A, \vec{x}_B)$. The use of these measures is subjective insofar as they depend on a specific domain or application. Conventionally, materials science studies begin by grouping similar materials in order to explore the patterns and rules in these materials. Consequently, measuring material similarity is considered a key technique in material informatics~\cite{bender2004molecular}. The advantages and disadvantages of many similarity measures were addressed in~\cite{maggiora2004molecular} and the argument that similar structures lead to similar properties was offered in~\cite{barbosa2004molecular, willett2014calculation}. However, the validity of this argument was reconsidered by Maggiora \textit{et al.}, who showed that small chemical modifications can lead to significant changes in biological activity~\cite{maggiora2013molecular}. Because the nature of materials is fundamentally diverse, Riniker \textit{et al.} addressed the problem of partially losing the transparency among fingerprint types by using fuzzier similarity methods~\cite{riniker2013similarity}. In addition, Maldonado \textit{et al.} optimized measures of molecular similarity and diversity based on selecting and classifying descriptors~\cite{maldonado2006molecular}. Moreover, several methods have been proposed for comparing crystalline materials~\cite{kupriyanov2015estimation,zhu2016fingerprint}.

Most similarity measures estimate the difference between two materials represented by vectors according to each vector’s dimension, and then provide an average for these differences. This can make the local differences (i.e., the difference in each dimension) fainter (or fuzzier). However, small modifications in materials can induce significant changes in the materials' properties, as mentioned in previous studies~\cite{maggiora2013molecular,riniker2013similarity}. This poses a key problem of how to select a similarity measure---that is, whether the loss in the materials' distinctiveness is acceptable when comparing materials in a specific context.

This study aims to establish the basis for choosing appropriate similarity measures between materials in a given context by bridging fundamental concepts in machine learning with the nature of material data. We focus on modeling materials' distinctiveness, and we explore whether an association exists between this property and the quality of approximating the energy function. By analyzing the characteristics of the energy function and descriptors, we propose novel quantitative protocols for selecting similarity measures.

\begin{figure*}
\centering
\includegraphics[scale=0.5]{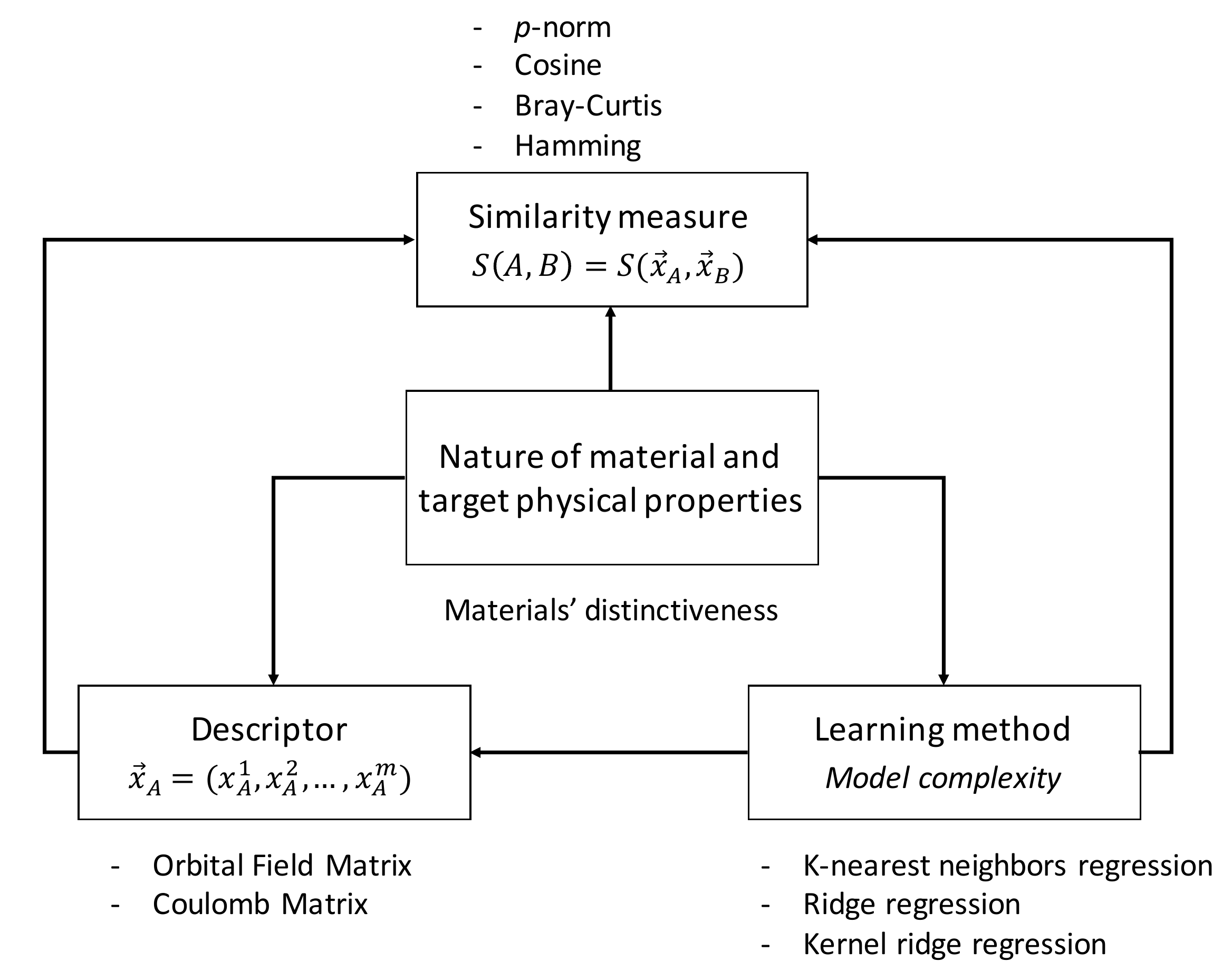}
\caption{Directed graph showing the dependence of selecting similarity measures on the descriptor and learning method on the one hand, and the dependence of selecting descriptors on the learning method, on the other hand, for effective materials data mining.}
\label{fig:methodology}
\end{figure*}

The paper is organized with four main sections, as follows:
\begin{itemize}
\item In Section~\ref{sec:descriptor}, we study the orbital field matrix and Coulomb matrix as descriptors to represent materials in vector space.  These descriptors can effectively predict materials' formation energies, as we explain based on previous studies.
\item In Section~\ref{sec:similarity}, we introduce several well-known similarity measures that are investigated in this study. In Section~\ref{sec:data}, we discuss the dataset used for this work.
\item In Subsection~\ref{sec:quantitative}, we propose a method for investigating how similarity measures of interest minimize the loss of materials' distinctiveness. From Subsection~\ref{sec:knn} to Subsection~\ref{sec:kernel_select}, by analyzing several learning methods that use similarity measures to predict crystal formation energies, we demonstrate that similarity measures need to be selected such that they fit with the characteristics of the descriptors and learning methods, as illustrated in Figure~\ref{fig:methodology}.
\end{itemize}
Our experimental results indicate that all of these methods show improved prediction performance when they capture the complexity of the energy function of crystals. Theoretical and empirical interpretations of these results from multiple perspectives reveal that descriptors that reflect the materials' distinctiveness (or identity) and similarity measures that minimize the loss of materials' distinctiveness help to improve the performance of formation energy prediction.

%In particular, we study the orbital field matrix and Coulomb matrix as descriptors to represent materials in vector space, which gave the effective prediction of materials' formation energies as presented in previous studies, in Section~\ref{sec:descriptor}. From Subsection~\ref{sec:knn} to Subsection~\ref{sec:kernel_select}, by analyzing several learning methods using similarity measures for predicting crystal formation energies, we demonstrate that similarity measures need to be selected such that they fit with the properties of the descriptors and learning methods, as illustrated in Figure~\ref{fig:methodology}. Our experimental results indicate that all of these methods show improved prediction performance when they capture the complexity of the energy function of crystals. Theoretical and empirical interpretations of these results from multiple perspectives reveal that descriptors that reflect the materials' distinctiveness (or identity) and similarity measures that minimize the loss of materials' distinctiveness help to improve the performance of formation energy prediction.

\section{Material descriptor}\label{sec:descriptor}
In this study, we aim to explore the rules of using similarity measures for materials by interpreting the formation energy prediction performance. Therefore, we consider two recently proposed vector representations of materials, which showed effective predictions of materials' formation energies: the Coulomb matrix (CM) and the orbital field matrix (OFM).

\subsection{Orbital field matrix}\label{sec:ofm}
The OFM is a novel descriptor that was proposed recently~\cite{lam2017machine,pham2018learning}. It uses the valence atomic configuration to represent the structure of materials. In the OFM representation, a material is assumed to be composed of building blocks that are called local structures. Each local structure includes a central atom and its environmental (neighboring) atoms. First, each atom is represented by a one-hot vector based on a dictionary of subshell orbitals: $D=\{s^1,s^2,p^1,...,p^6,d^1,...,d^{10},f^1,...,f^{14}\}$. We denote the vector of the central atom by $\vec O_{central}$, and the vector of the $k^{th}$ neighboring atom by $\vec O_k$. Second, the vector representing the environment of each atom in a structure, $\vec O_{env}$, is computed as follows:
\begin{equation}
\vec O_{env} = \sum_k^{K} w_k \vec{O}_k,
\end{equation}
where the weight, $w_k$, measures the contribution of the $k^{th}$ neighboring atom, and $K$ is the number of neighboring atoms. The local structure is represented by a matrix, $X$, where $X_{ij}$ represents the number of an environmental atomic orbital (orbital $j$) coordinated with a central atomic orbital (orbital $i$). Hence, the representation matrix of a local structure is
\begin{equation}
\begin{aligned}
X &= \vec{O}^T_{central} \times \vec{O}_{env} \\
&= \vec{O}^T_{central} \times \Big( \sum_k^K \vec{O}_k \frac{\theta_k}{\theta_{max}} \Big)
\end{aligned}
\end{equation}
where $w_k = \frac{\theta_k}{\theta_{max}}$ is the weight representing the contribution from atom $k^{th}$ to the coordination number of the central atom; $\theta_k$ is the solid angle determined by the face of the Voronoi polyhedral that separates the $k^{th}$ atom and the central atom; and $\theta_{max}$ is the maximum of all solid angles determined by this Voronoi polyhedral.

The distance $r_k$ between the central atom and the $k^{th}$ neighboring atom is incorporated in the representation of local structures as follows:
\begin{equation}
X = \vec{O}_{central}^T \times \Big( \sum_k^K \vec{O}_k \frac{\theta_k}{\theta_{max}} \zeta(r_k) \Big),
\end{equation}
where $\zeta(r_k) = 1/r_k$ is the distance-dependent weight function. Finally, the descriptor for the entire material is a mean of descriptors for its local structures. 
%\begin{equation}
%X^{(p)}_{ij} = \sum_{k\in n_p}o_j^ko_i^{(p)}\frac{\theta_k^{(p)}}{\theta_{max}^{(p)}}\zeta(r_{pk}),
%\end{equation}
%where $\zeta(r_{pk}) = 1/r_{pk}$ is the distance-dependent weight function. The descriptor for entire material here is a mean of descriptors for its local structures.

In an extension of the OFM, the information regarding the central atom is incorporated by simply concatenating $\vec{O}^T_{central}$ to the matrix $X$ as a new column, as follows:
\begin{equation}
X = \vec{O}^T_{central} \times \Big( 1.0, \sum_{k}^{K} \vec{O}_k \frac{\theta_k}{\theta_{max}}\zeta{(r_k)} \Big)
\end{equation}
In this study, we use this extension to the OFM to predict crystals' formation energies.

\subsection{Coulomb matrix}\label{sec:cm}
The CM~\cite{rupp2012fast,montavon2012learning} is a descriptor that encodes the structure of a material using nuclear charges $Z_i$ and the 3D coordinates $\textbf{R}_i$ of each constituent atom in the material, as follows:

%\[
%C_{ij} = 
%	\begin{cases}
%		0.5Z_i^{2.4} &\quad \forall i=j \\
%		\frac{Z_iZ_j}{|\textbf{R}_i - \textbf{R}_j|} &\quad \forall i \neq j
%	\end{cases}
%\]
\begin{equation}
C_{ij} = 
\begin{cases}
0.5Z_i^{2.4} &\quad \forall i=j \\
\frac{Z_iZ_j}{|\textbf{R}_i - \textbf{R}_j|} &\quad \forall i \neq j
\end{cases}
\end{equation}
To deal with the atom-ordering problem in CM, the authors used (i) the eigenspectrum representation that first obtains eigenvalues of each Coulomb matrix, and then uses the sorted eigenvalues (i.e., spectrum) as the representation, and (ii) sorted Coulomb matrices that choose the permutation of atoms whose associated Coulomb matrix $C$ satisfies $||C_i|| \geq ||C_{i+1}|| $ $\forall i$, where $C_i$ is the $i^{th}$ row of the Coulomb matrix. In practice, padding the Coulomb matrices by zero-valued entries is required in order to avoid the difference in matrix size induced by the difference in the number of atoms in each material.

\section{Similarity measures of interest}\label{sec:similarity}
%In materials informatics, the similarity between materials is quantified through several measures. These measures are called distances if they satisfy all conditions of a metric (i.e., non-negativity, identity of indiscernibles, symmetry, and triangular inequality). If they do not fully satisfy these conditions, they are called dissimilarity measures.
In materials informatics, the similarity between materials is quantified through several measures that mostly estimate the difference between materials. These measures are called distances if they satisfy all conditions of a metric (i.e., non-negativity, identity of indiscernibles, symmetry, and triangular inequality). If they do not fully satisfy these conditions, they are called dissimilarity measures.

Let $u,v \in \mathbb{R}^m$ be two vectors. Similarity measures are mathematical functions taking $u$ and $v$ as their input with a scalar as output. In this study, we investigate several well-known similarity measures that are commonly utilized in the vector space, as follows:
\begin{itemize}
\item Hamming distance:  $d(u,v) = {\sum_{i=1}^m[u_i \neq v_i]}/{m}$, the Hamming distance between two vectors here is computed by counting the number of different elements in these vectors. With OFM and CM, material vectors are numerical vectors, but they share common zero-valued elements. Hence, their Hamming distance may differ from one.
\item $p$-norm distance: $d(u,v) = ||u-v||_p = (\sum_{i=1}^{m}|u_i-v_i|^p)^{\frac{1}{p}}$ with $p=1,2,3$ in which the 1-norm and 2-norm are known as the Manhattan and Euclidean distances, respectively.
\item Cosine distance: $d(u,v) = 1 - u.v/||u||_2||v||_2$.
\item Bray-Curtis (B-C) dissimilarity: this is not a distance measure because it does not obey the triangular inequality, $d(u,v) = \sum_{i=1}^{m}|u_i - v_i|/\sum_{i=1}^{m}|u_i + v_i|$
\end{itemize}

\section{Data}\label{sec:data}
This study utilizes the dataset of magnetic materials based on rare earth-transition metal alloys extracted from the Open Quantum Materials Database (OQMD)~\cite{kirklin2015open, saal2013materials}. The dataset consists of 5967 crystals. Crystals containing rare-earth and transition elements are considered because the diversity of their structures induces a diverse range of electronic properties on account of the interval magnetic freedom~\cite{hinatsu2015diverse,lynch1987effect}. In other words, the distinctiveness of crystals and their properties as well as the importance of considering small changes in crystals are the main characteristics of this dataset and thus useful for our study.

\section{Similarity measure selection based on an analysis on descriptors and model complexity}

\subsection{Quantitative evaluation of the materials' distinctiveness loss using similarity measure}\label{sec:quantitative}
%The material distinctiveness requires considering small differences between materials by measuring their dissimilarity. This entails averaging local differences insofar as most distances and dissimilarity measures used in the vector space lead to the loss of material identity (or distinctiveness). In dissimilarity measures, the Hamming distance measures the minimum number of substitutions required to change one vector to the other, and preserves the difference in each vectors' dimension. Therefore, the Hamming distance maximally preserves the material identity.

Most similarity measures used in the vector space take an average of the difference in each vector's dimension, leading to the loss of the materials' distinctiveness. To preserve the materials' distinctiveness when measuring the material similarity we should accumulate all the differences in every vector's dimension. The Hamming distance, which counts the minimum number of substitutions required to change one vector to the other, can be considered the simplest measure that fits this purpose. 

In fact, totally preserving the materials' distinctiveness is meaningless in terms of discovering knowledge because it prevents the generalization of information when seeking patterns or rules. Therefore, the loss of the distinctiveness between materials can be tolerated. However, the extent to which this is tolerated must be adjusted based on the nature of data and the specific purposes when mining. To estimate the loss of materials' distinctiveness when using similarity measures, we compute the correlation between the pairwise similarity of materials created by these measures and that created by the Hamming distance. In other words, given a measure, if two materials are different (or if they are distant in the vector space), this is determined by both this measure and the Hamming distance. This measure will be considered to preserve the materials' distinctiveness.

We estimate affinity matrices of materials represented by OFM and CM with the Hamming distance, the $p-$norm distance, the cosine distance, and the B-C dissimilarity. Next, we calculate the correlation coefficients between flattened forms of these matrices, as shown in Table~\ref{tab:correlation}. The table shows that the 1-norm (Manhattan) distance and B-C dissimilarity correlate more to the Hamming distance than the others with both descriptors. Therefore, the use of these two measures results in only a small loss of the materials' distinctiveness.

\begin{table}
\centering
%\caption{Correlation of affinity matrices created by the $p$-norm distance, cosine distance, and B-C dissimilarity towards the affinity matrix created by the Hamming distance}
\caption{Estimation of the loss of materials' distinctiveness when using similarity measures by estimating the correlation of affinity matrices created by the $p$-norm distance, cosine distance, and B-C dissimilarity towards the affinity matrix created by the Hamming distance.}
\label{tab:correlation}
\begin{tabular}{c|c|c} \hline
Descriptor& Dissimilarity measure & Correlation\\ \hline
\multirow{5}{*}{OFM} & 1-norm & \textbf{0.624}\\
& 2-norm  & 0.45\\
& 3-norm & 0.37\\
& cosine & -0.569\\
& B-C & \textbf{0.564}\\ \hline
\multirow{5}{*}{CM} & 1-norm & \textbf{0.57} \\
& 2-norm & 0.394\\
& 3-norm & 0.345\\
& cosine & -0.308\\
& B-C & \textbf{0.551}\\ \hline
\end{tabular}
\end{table}

\subsection{K-nearest neighbors regression}\label{sec:knn}
K-nearest neighbors (KNN) is known as a ``lazy learning'' algorithm. It predicts a target value of an instance by averaging nearest neighbor target values of this instance without any assumption regarding the relation between this instance and its target value. As such, KNN is useful when exploring the nature of data because real-world data does not obey any typical theoretical assumption. In KNN, there are two hyperparameters that need to be defined beforehand: (i) the number of nearest neighbors, denoted by $k$, and (ii) an appropriate similarity measure between instances. In the case study predicting crystal formation energy, we aim to clarify the importance of taking descriptors and the number of nearest neighbors into account when selecting similarity measures. 

Let $\mathcal{D} = \{(x_1, y_1), (x_2, y_2),..., (x_n, y_n)\}$ denote sample data generated from a function $y=f(x)$. For a new instance $x'$, KNN estimates the function value for this instance as
\begin{equation}
\hat{f}(x') = \frac{1}{|N_k|} \sum_{(x_i, y_i) \in N_k}y_i
\end{equation}
where $N_k \subset \mathcal{D}$ is the set of $k$ nearest neighbors of $x'$.

In KNN, we do not extract generalized patterns or models from instances in $\mathcal{D}$. This method utilizes all data points to approximate the function $f(x)$. The model structure is characterized by the number of nearest neighbors ($k$) and the similarity measure between instances.

Owing to a lack of generalization, overfitting can occur in KNN, particularly when $k$ is too small. Overfitting occurs when the model aptly predicts instances in the existing data but poorly predicts new instances. Therefore, in some cases, the high prediction performance of KNN is irrelevant when exploring the nature of data because of its regularization. It tolerates prediction errors in existing data in order to better predict new instances. In KNN, using a large $k$ value reduces the risk of overfitting. However, a large $k$ may not be a suitable approximation for the existing data if the function $f(x)$ is a complex curve with many extreme points, as illustrated in Appendix A.

This study focuses on exploring the nature of a given material dataset by interpreting empirical prediction results. Thus, overfitting must be avoided to prevent confusion in the interpretation. If the data is generated from multiple distributions, we separately fit the model for each group of instances assumed to be generated from a distribution.

To effectively approximate the energy function, we need to understand characteristics of this function. Visualizing the energy function is a simple way to derive an intuitive understanding of this function. This process is discussed in Subsection~\ref{sec:visualization}. In Subsection~\ref{sec:knn_prediction}, we present the experimental results from predicting crystals' formation energies using KNN, and we offer several remarks on the number of neighbors and similarity measures. Our interpretation of the number of neighbors, used to fit the energy function, reveals the complexity of this function. The details of this interpretation are presented in Subsection~\ref{sec:optimalk}. Relying on the characteristics of the energy function and descriptors, we can select appropriate similarity measures to obtain high formation energy prediction performance, as described in Subsections~\ref{sec:knn_sim_energy} and~\ref{sec:desciptor_identity}.

\subsubsection{Formation energy surface visualization}
\label{sec:visualization}
\begin{figure*}
\centering
%\subfigure{\includegraphics[scale=0.4]{fig_energySurface1.pdf}}
\subfigure{\includegraphics[scale=0.3]{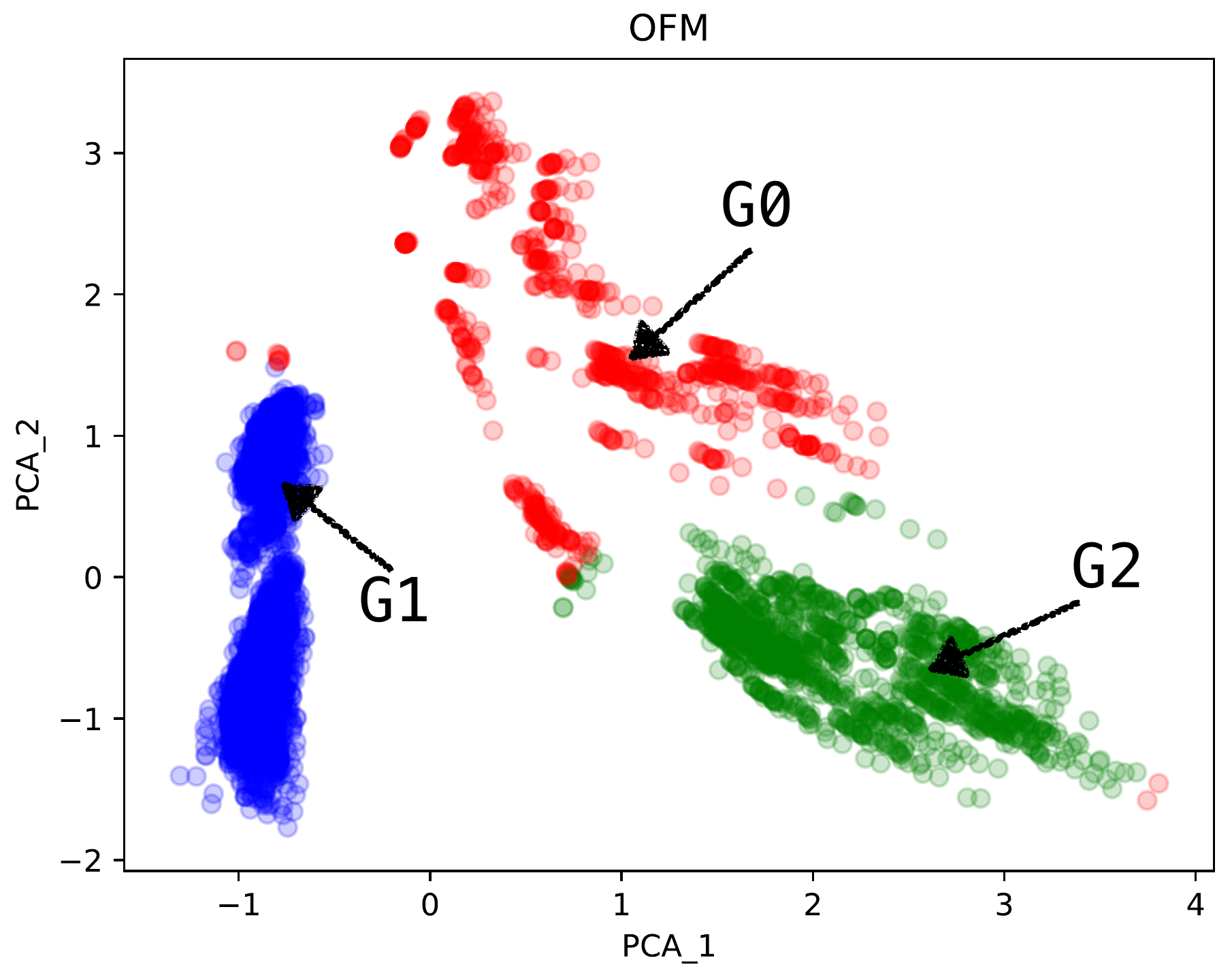}}
\subfigure{\includegraphics[scale=0.3]{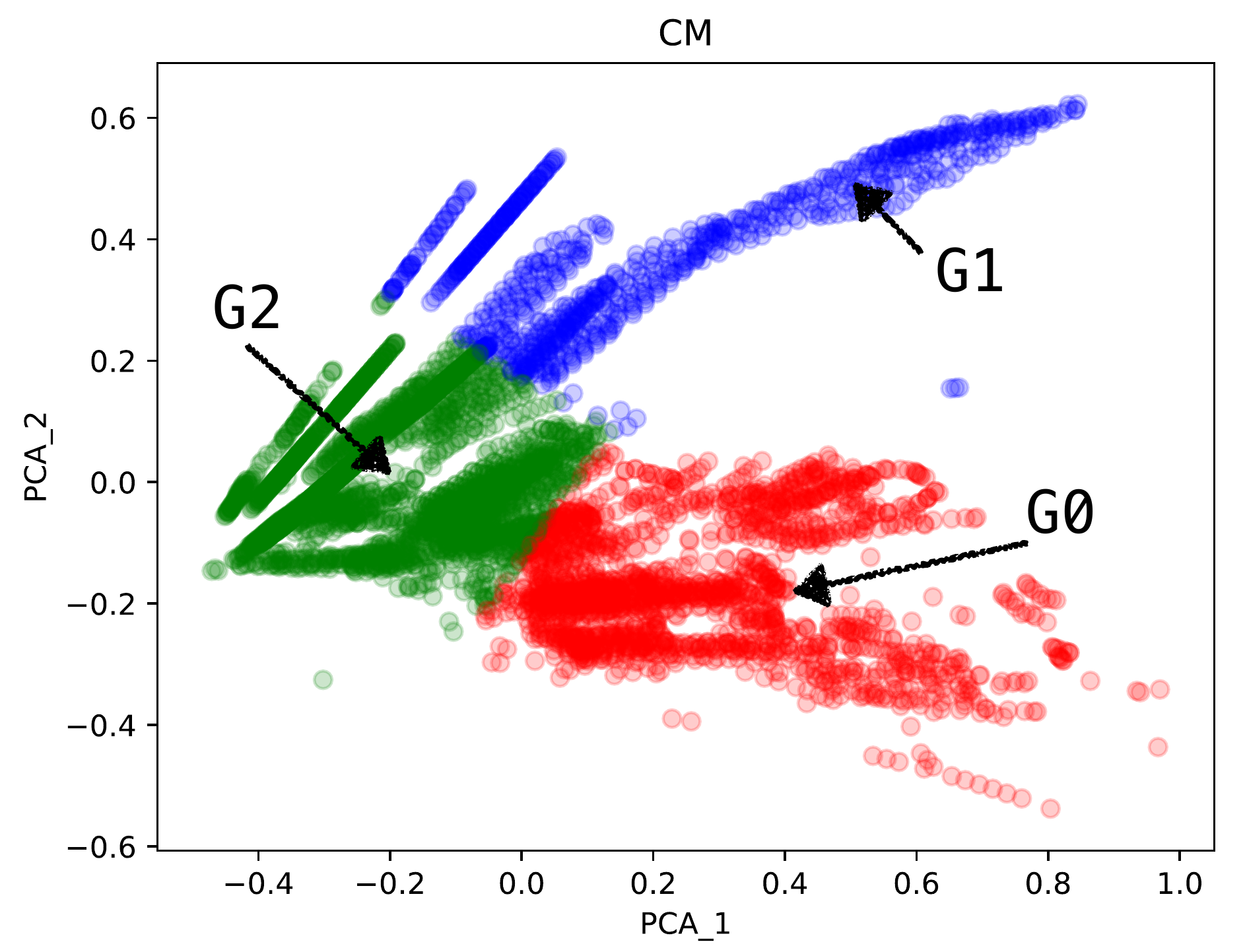}} \\
%\subfigure{\includegraphics[scale=0.4]{fig_energySurface2.pdf}}
\subfigure{\includegraphics[scale=0.5]{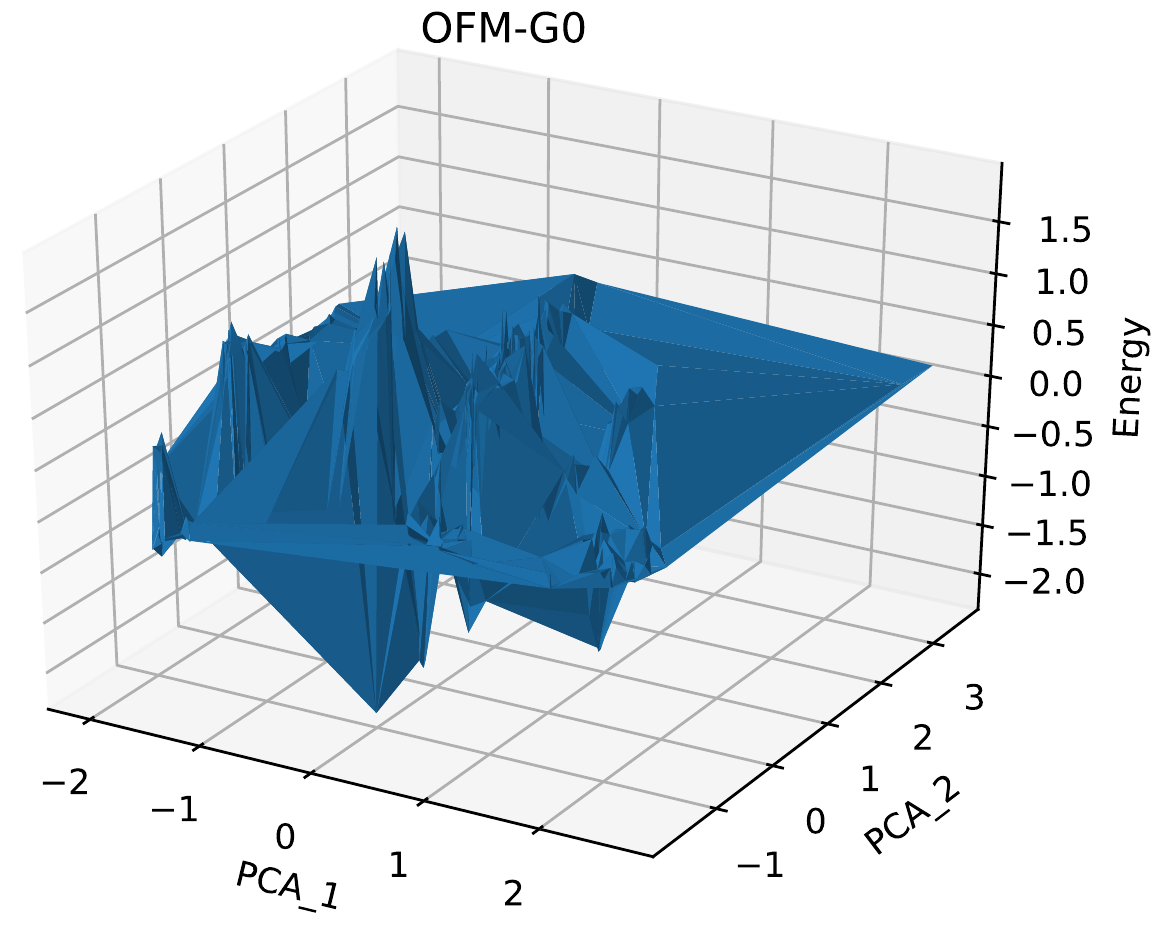}}
\subfigure{\includegraphics[scale=0.3]{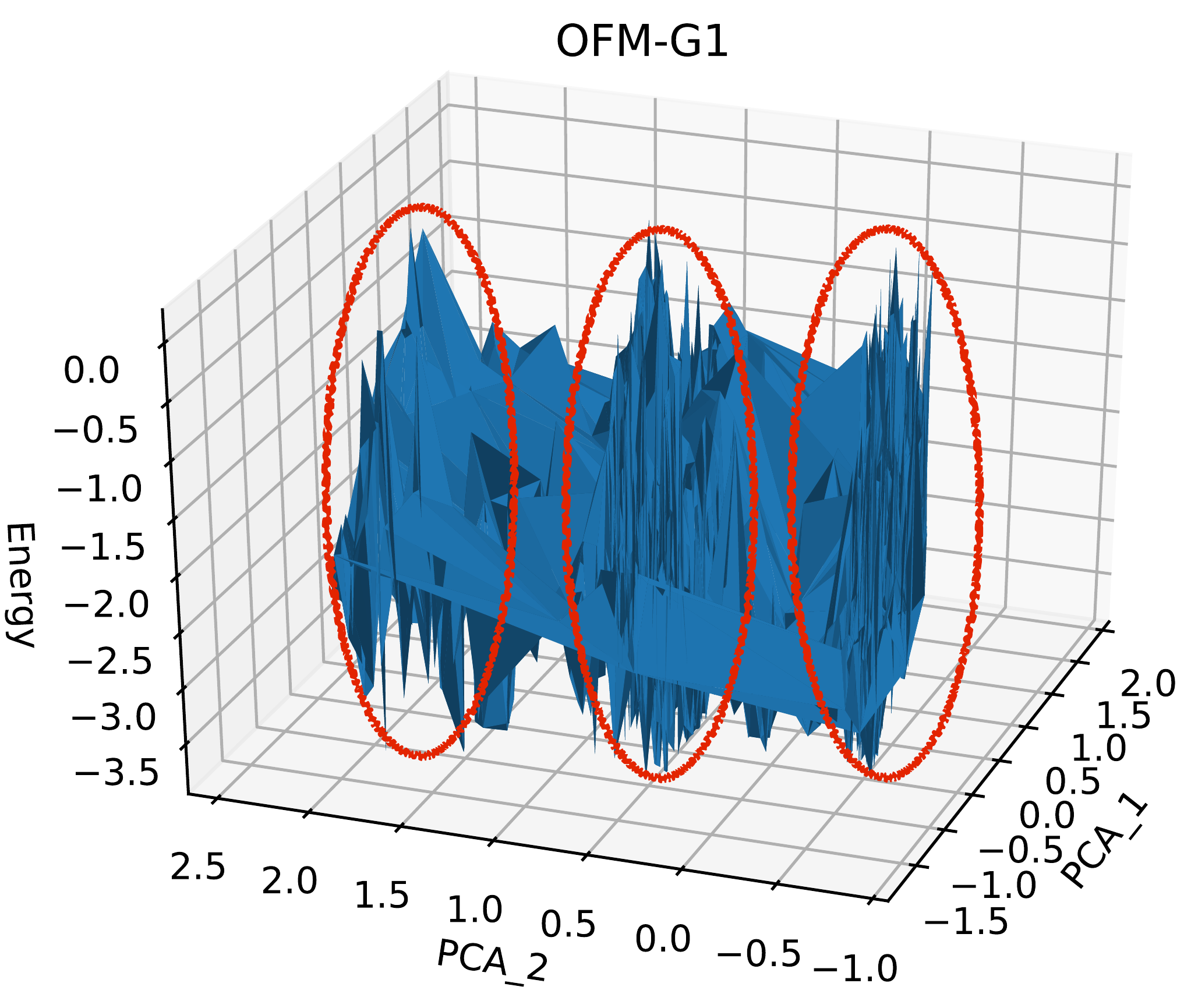}}
\subfigure{\includegraphics[scale=0.5]{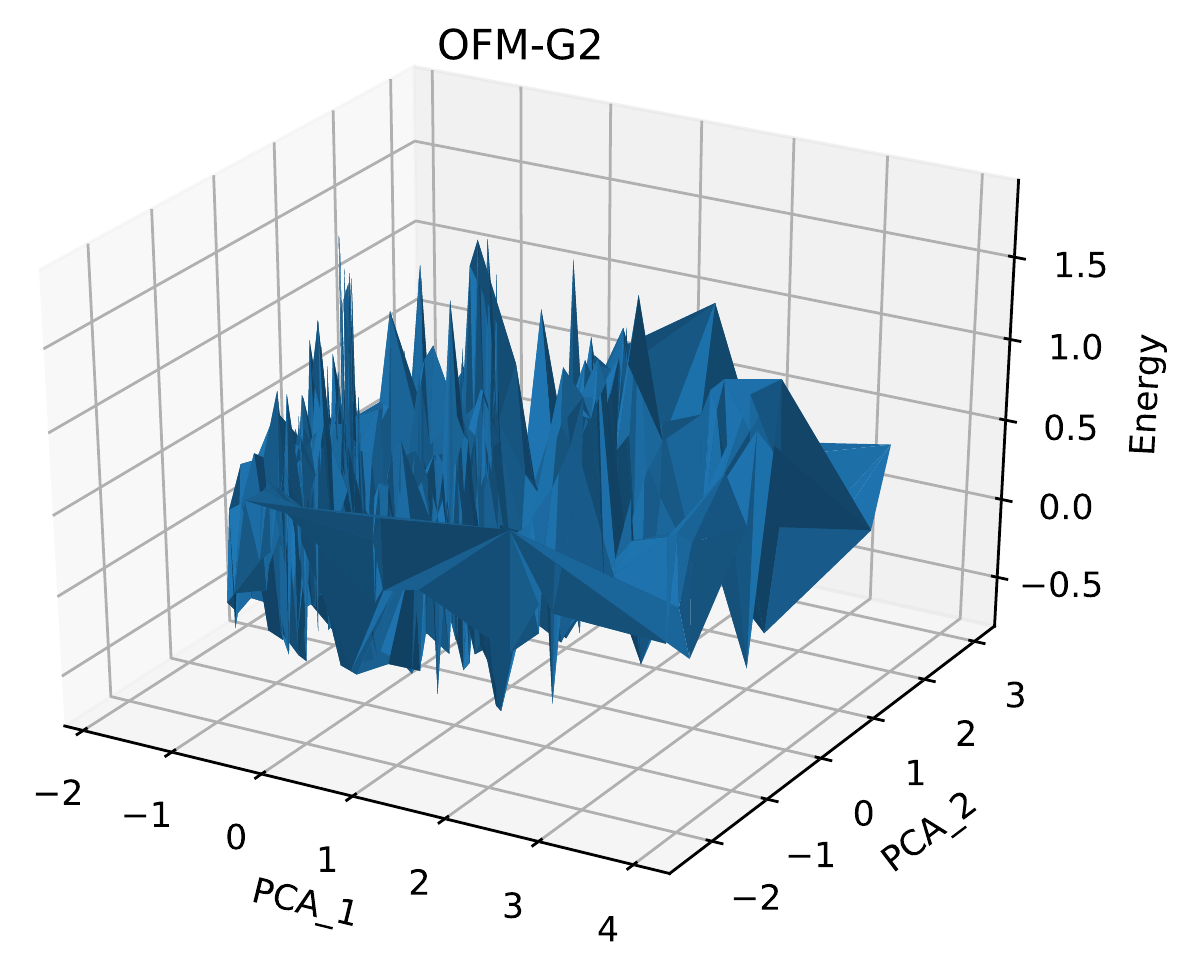}}
\subfigure{\includegraphics[scale=0.5]{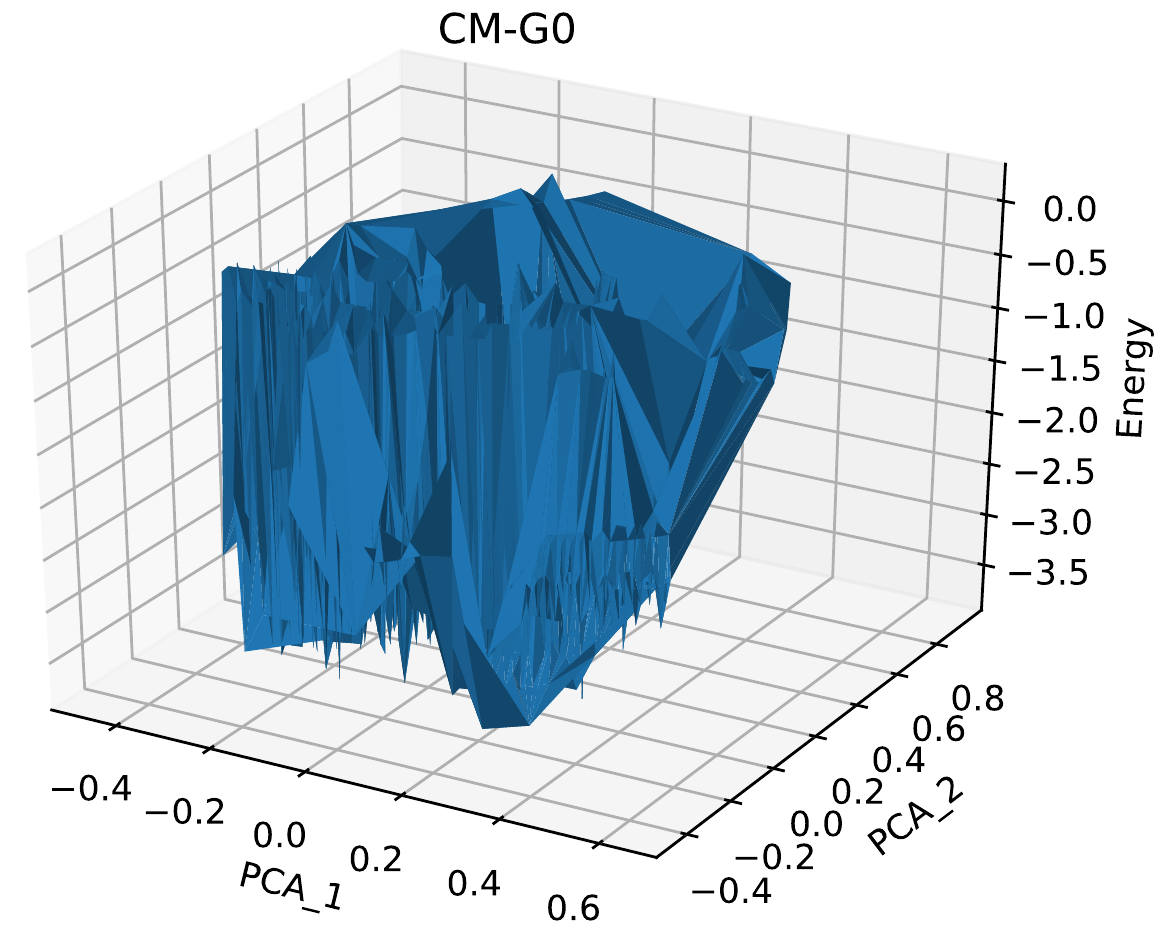}}
\subfigure{\includegraphics[scale=0.5]{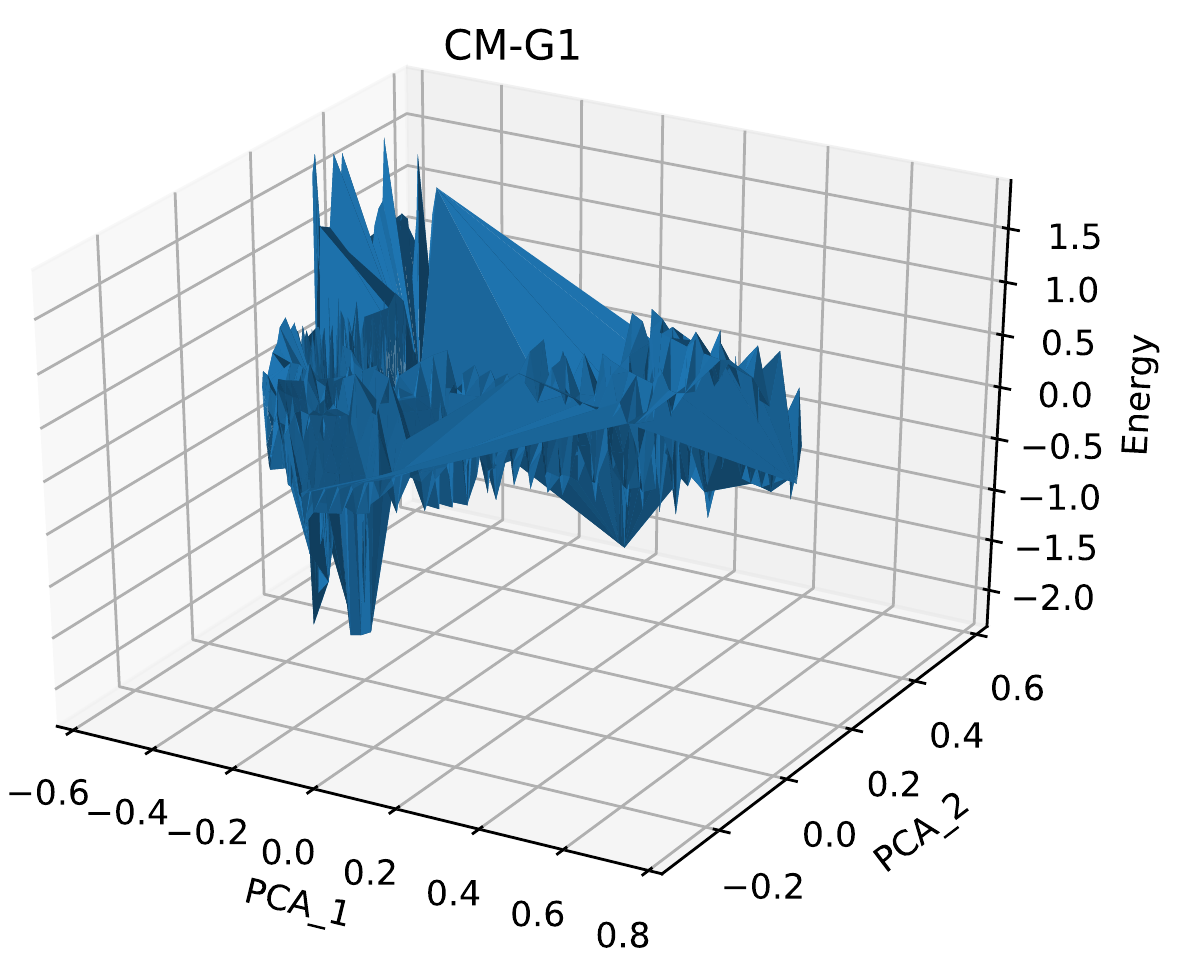}}
\subfigure{\includegraphics[scale=0.5]{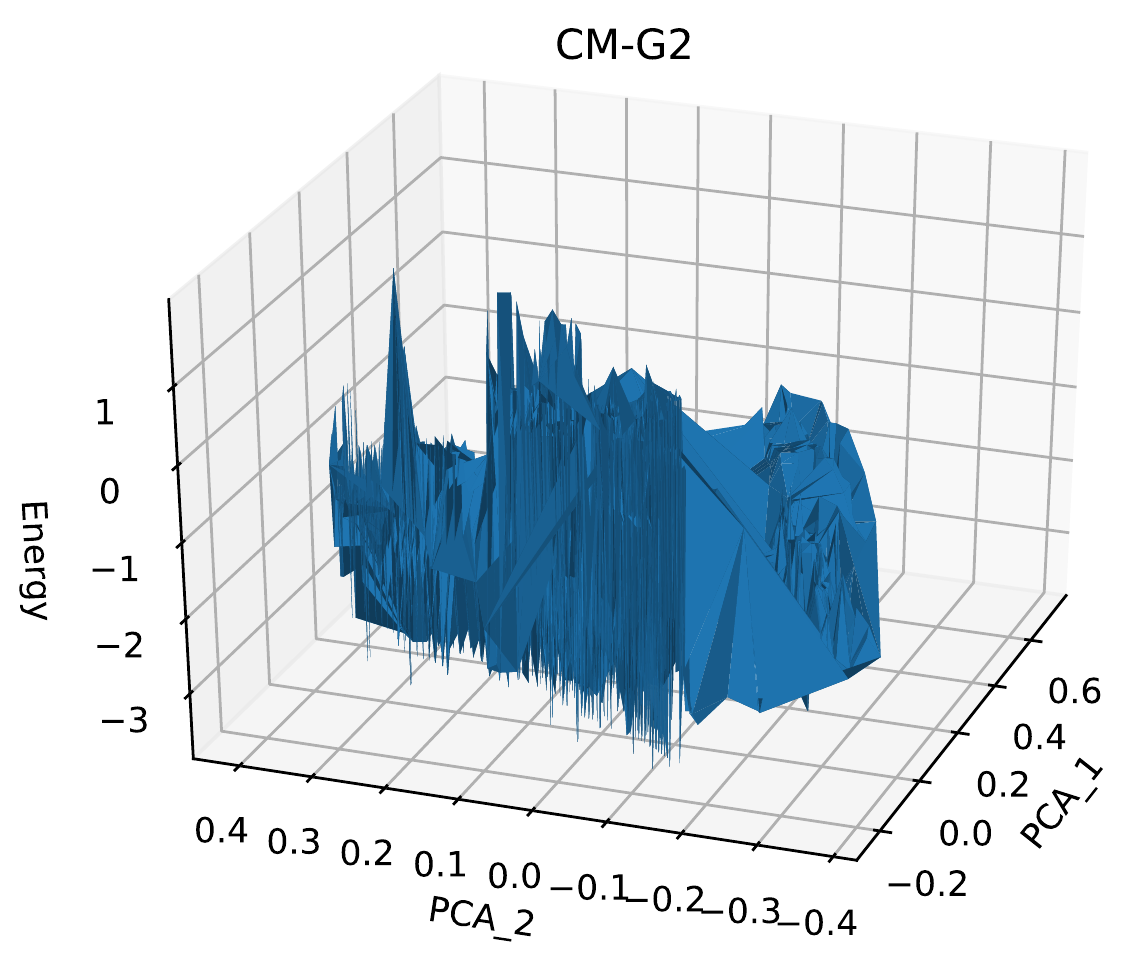}}
\caption{Illustration of the formation energy surface over crystals in six samples: CM-G0, CM-G1, CM-G2, OFM-G0, OFM-G1, and OFM-G2. The groups of data points in the dataset are in red, blue, and green. The sample OFM-G1 shows its three sub-groups (bounded by red circles) inside.}
\label{fig:energy}
\end{figure*}

In this study, $f(x)$ is the energy function of crystals. Visualizing the energy surface of crystals can help with a preliminary and intuitive understanding of the properties of this function. To visualize this, we use principal component analysis (PCA) to project data instances to a 2D subspace. Next, we plot the energy surface with the projected data in Figure~\ref{fig:energy}.

Figure~\ref{fig:energy} shows that the data is diverse because it includes several groups of instances under different energy functions. To avoid overfitting, we separate the data into groups, and considered each group individually. We divide the dataset (comprising 5967 crystals) into three groups to make the number of instances in each group sufficiently large to train the model. After clustering, we obtain six samples, denoted by CM-G0, CM-G1, CM-G2, OFM-G0, OFM-G1, and OFM-G2. The energy surface of each sample is also plotted, as shown in Figure~\ref{fig:energy}. The visualization shows that the energy surface in all samples is complex, with many extreme points in their small vicinities.

\subsubsection{Remarks from experimental results}\label{sec:knn_prediction}
\begin{figure*}
\centering
\subfigure{\includegraphics[scale=0.5]{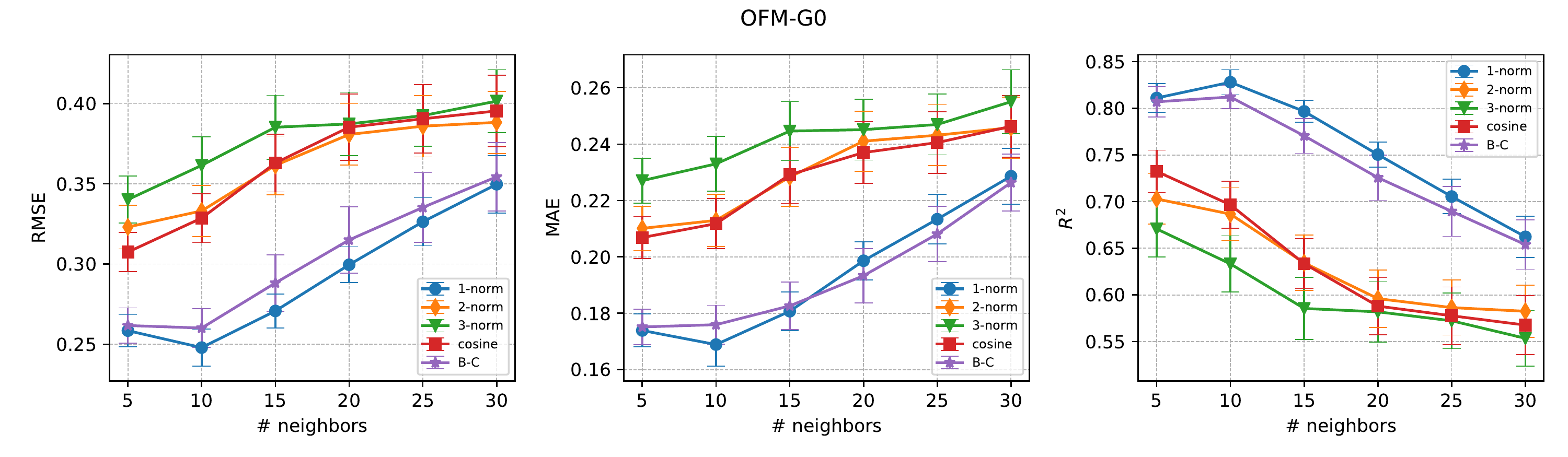}}
\subfigure{\includegraphics[scale=0.5]{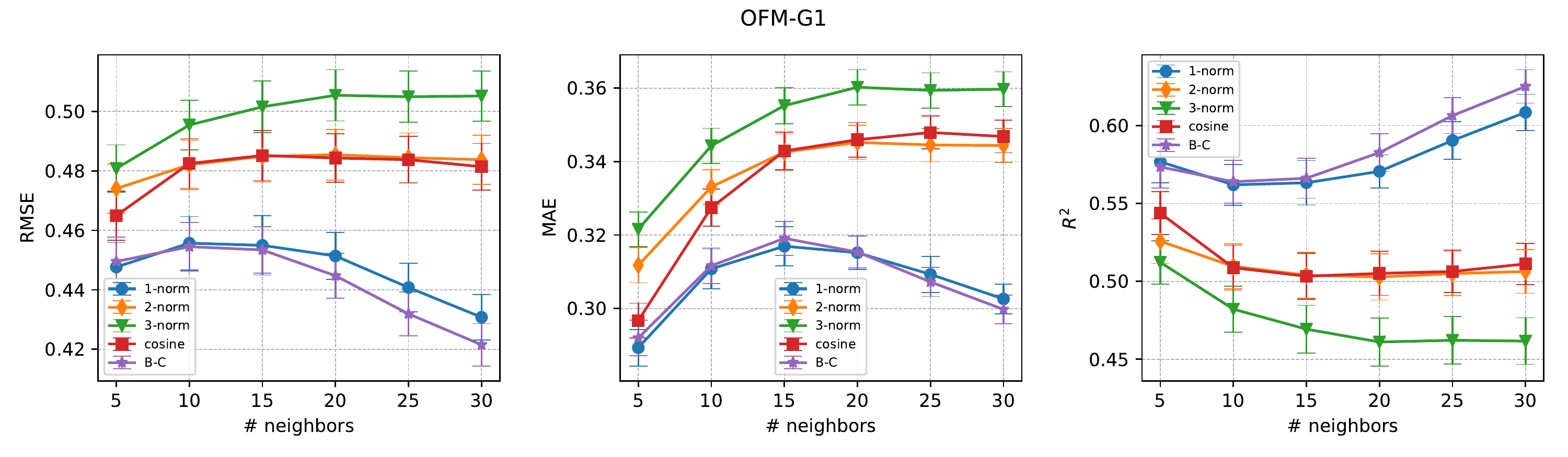}}
\subfigure{\includegraphics[scale=0.5]{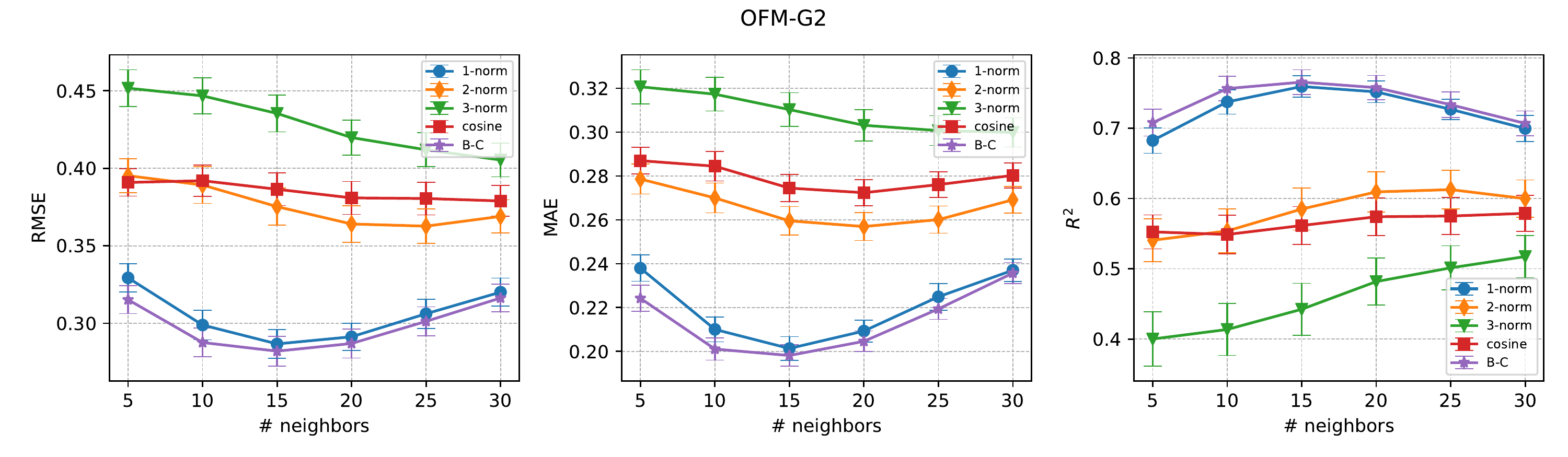}}
\caption{Crystal formation energy prediction performance using KNN with various values of $k$ and similarity measures for samples OFM-G0, OFM-G1, and OFM-G2 (using the OFM descriptor).}
\label{fig:knn-acc-ofm}
\end{figure*}

\begin{figure*}
\centering
\subfigure{\includegraphics[scale=0.5]{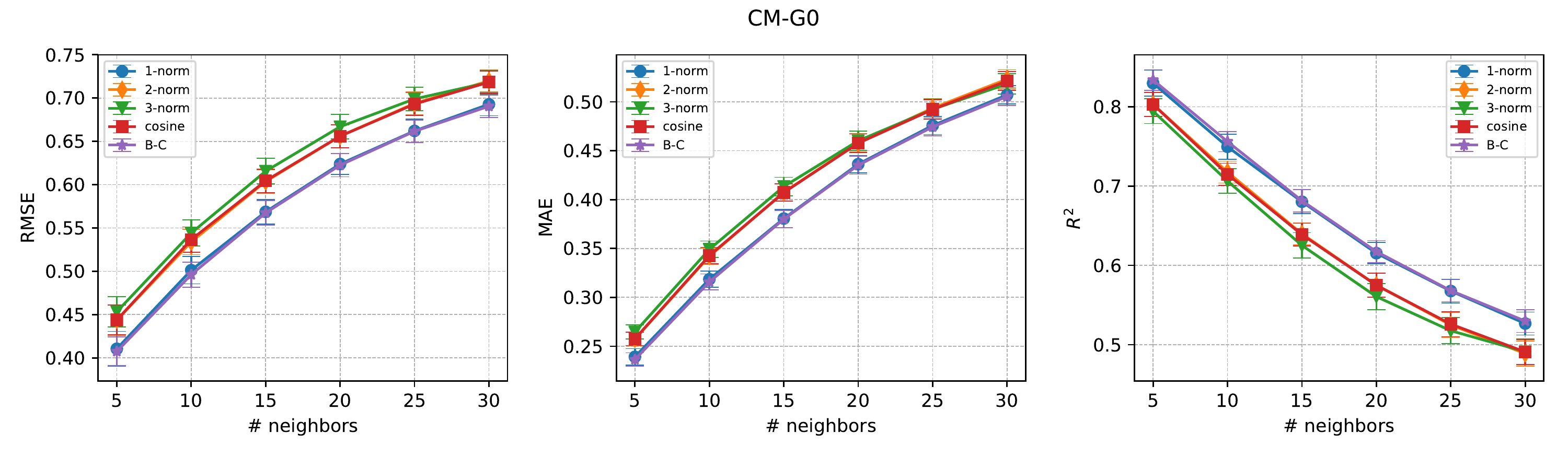}}
\subfigure{\includegraphics[scale=0.5]{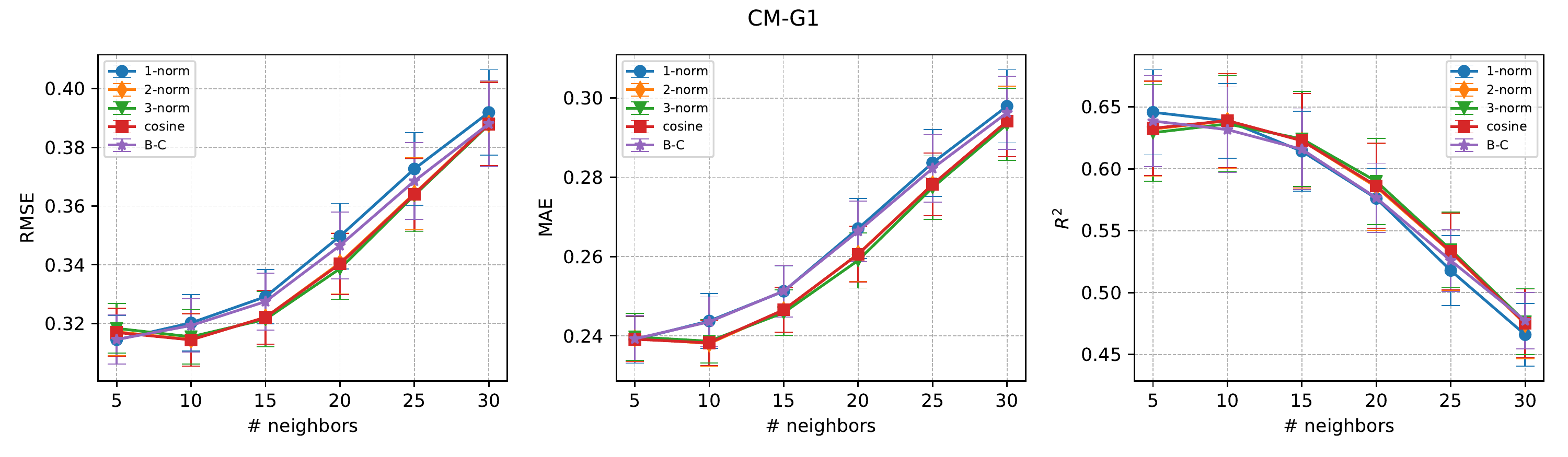}}
\subfigure{\includegraphics[scale=0.5]{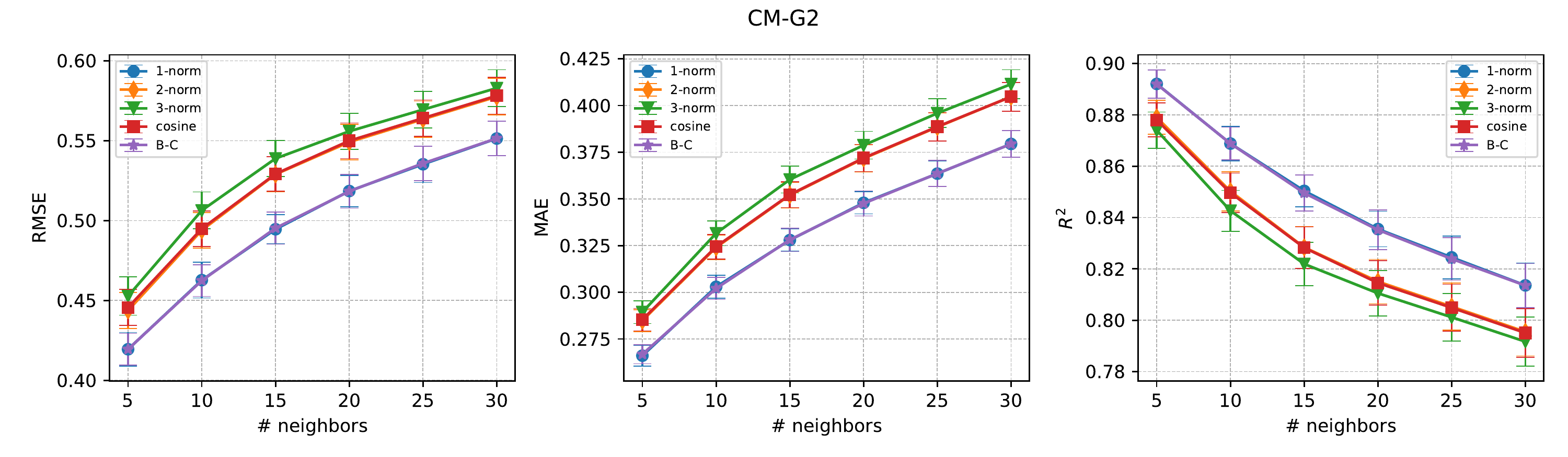}}
\caption{Crystal formation energy prediction performance using KNN with various values of $k$ and similarity measures for samples CM-G0, CM-G1, and CM-G2 (using the CM descriptor).}
\label{fig:knn-acc-cm}
\end{figure*}

To find the most appropriate value for $k$ and a similarity measure for different representations of the data, we perform ten-fold cross-validation with OFM and CM for six samples, as shown in Figures~\ref{fig:knn-acc-ofm},~\ref{fig:knn-acc-cm}. The prediction performance is evaluated using the root mean squared error (RMSE), mean absolute error (MAE), and $R^2$ (the coefficient of determination).

The experimental results show that a small $k$ ($k=5$ or $k=10$) results in the highest accuracy for most samples. In addition, increasing the value of $k$ degrades the prediction performance. As an exception, in sample OFM-G2, $k=15$ results in better performance than with smaller values of $k$. In sample OFM-G1, larger $k$ values result in better performance in terms of the RMSE and $R^2$.

The 1-norm distance and B-C dissimilarity, which minimize the loss of materials' distinctiveness, as explained in Section~\ref{sec:quantitative}, result in more accuracy than other measures for most samples. With these two measures, we compare two descriptors and found that using these measures for OFM results in more improvement than using them for CM. In addition, the 1-norm distance and B-C dissimilarity are worse than the others in CM-G1. This shows the dependency of choosing similarity measures for the descriptors.

\subsubsection{The optimal value of $k$ reveals the complexity of the energy surface}
\label{sec:optimalk}
%Figure~\ref{fig:energy} shows that the energy surface is flexible and complex with many extreme points within a small vicinity.
As mentioned above, KNN locally approximates $f(x)$ by averaging the energies of the nearest neighbors of each data point. For a data point $(x_i, y_i)$, let $N_k^+ \subset N_k$ be a subset of neighbors in $k$ nearest neighbors whose target values are greater than $y_i$: $y_j = y_i + \delta_j$ with $\delta_j > 0$. Let $N_k^- \subset N_k$ be the subset of neighbors whose target values are smaller than $y_i$: $y_j = y_i - \delta_j$ with $\delta_j > 0$. We note that $|N_k^+| + |N_k^-| = |N_k|$ . The formula for estimating the target value of $x_i$ is re-written as follows: 

\begin{equation}\label{eq:average}
\hat{f}(x_i) = \frac{1}{|N_k|} \times \Bigg( \sum_{(x_j, y_j) \in N_k^+} (y_i + \delta_j)+  \sum_{(x_j, y_j) \in N_k^-} (y_i - \delta_j) \Bigg)
\end{equation}

Relying on Equation~\ref{eq:average} to precisely estimate $\hat{f}(x_i)$ requires that neither $N_k^+$ nor $N_k^-$ are empty and that the positive residual $\sum_{(x_j, y_j) \in N_k^+}\delta_j$ and the negative one $-\sum_{(x_j, y_j) \in N_k^-}\delta_j$ can eliminate each other. Therefore, instances, which are extreme points in the energy function (see Appendix A), are difficult to fit using KNN. The nearest neighbors of these extreme points often belong in their vicinity, and energies of these neighbors ($y_j$) are only smaller or greater than $y_i$ (i.e., $N_k^+$ or $N_k^-$ is empty). Thus, the use of a small number of neighbors can help to reduce the residual in the estimation in such a situation. Therefore, a small $k$ value, which is optimal for most of the samples (as shown in Figures~\ref{fig:knn-acc-ofm},~\ref{fig:knn-acc-cm}), is consistent with the fact that the energy surface is rough, with many extreme points within a small vicinity, as shown in Figure~\ref{fig:energy}. 

The experimental results show that large values of $k$ perform better in sample OFM-G1. The underlying reason for this is that the sample can be divided into three smaller groups under three different functions, as bounded by the red circles in Figure~\ref{fig:energy}. Thus, choosing a small $k$ value can lead to overfitting in this sample and degrade the prediction performance. In this case, using a large $k$ plays the role of regularization when dealing with overfitting. Therefore, using a large $k$ in sample OFM-G1 does not conflict with our hypothesis regarding the relation between $k$ and the complexity of the energy function.

\subsubsection{Similarity measure selection based on the energy function complexity}\label{sec:knn_sim_energy}
\begin{figure}
\centering
\includegraphics[scale=0.4]{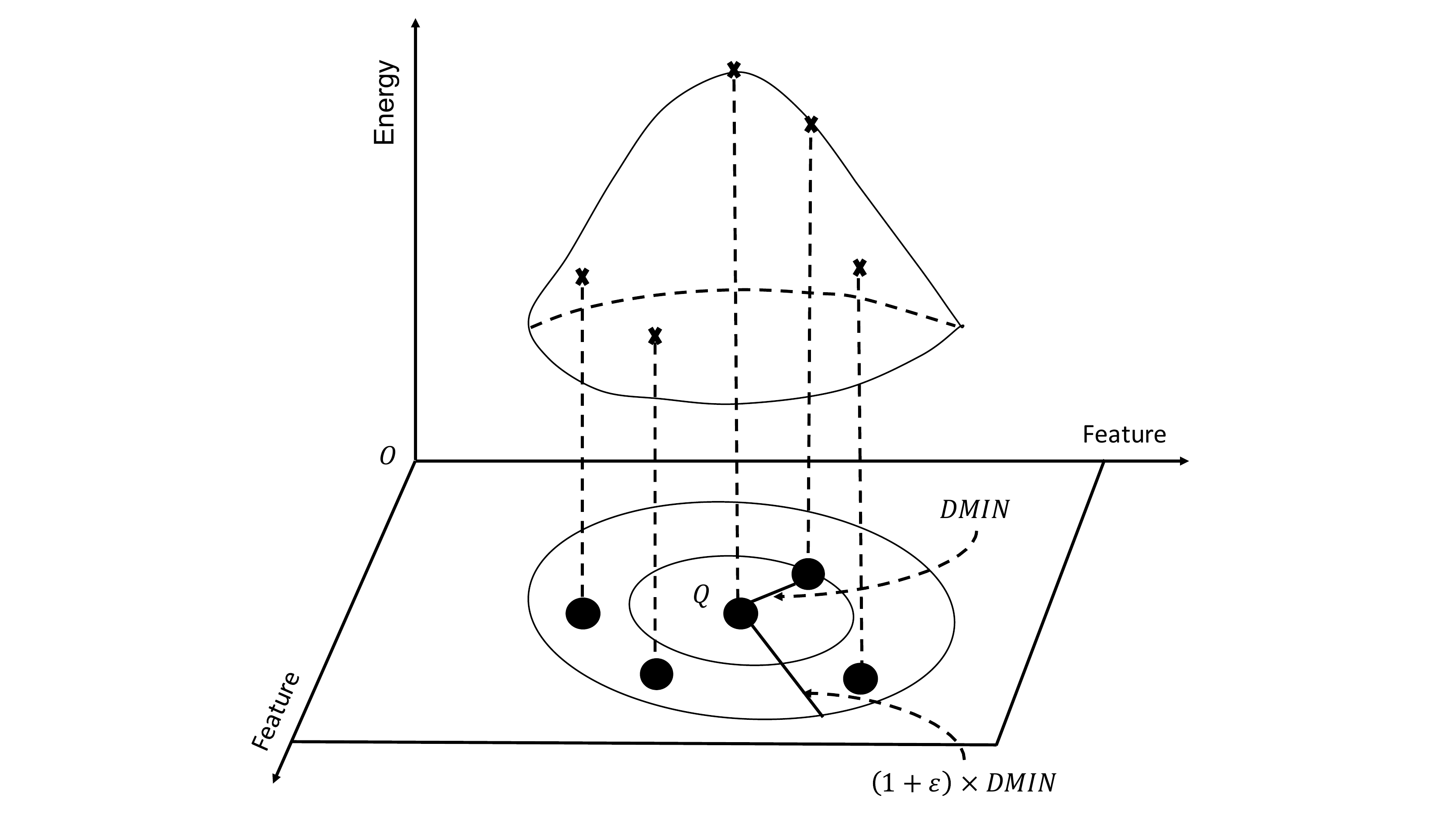}
\caption{Investigation of similarity measures in KNN. Each query point $Q$ is enlarged by a neighboring region. If other data points belong to this region, they are considered neighbors of the query point. The number of data points in the neighboring region of the query point depends on the similarity measure utilized.}
\label{fig:similarityfitextreme}
\end{figure}

The analysis of the value of $k$ presented above reveals the complexity of the energy function used as the basis for investigating similarity measures. Suppose that $Q$ is an extreme point of the function, and that we need to approximate the energy value at this point. $Q$ is called a query point. The closest point to the query point is determined, and the distance between these points is denoted by $DMIN$. To identify other neighbors of $Q$, we enlarge the region surrounding $Q$ by a radius $(1 + \varepsilon) \times DMIN$, called the neighboring region of $Q$. Data points within this region are considered neighbors of the query point (in Figure~\ref{fig:similarityfitextreme}).

Alternatively, rather than determining $k$ nearest neighbors, KNN can take an average of all neighbors belonging to the neighboring region of each point in the data (query point), determined by a distance threshold. This method is called fixed-radius nearest-neighbors regression~\cite{chen2018explaining}. To estimate the energy at $Q$, we average all data points falling in its neighboring region. Of course, the number of neighbors of $Q$ depends on which similarity measure is used. As mentioned above, to predict the energy at $Q$ precisely, we need to select similarity measures that help to identify a small number of neighbors in the vicinity of $Q$. Although different measures will produce values in different ranges, they share the common factor $\varepsilon$. Thus, $1 + \varepsilon$ can be understood as the relative value of these measures. As such, it is possible to compare these measures.

\begin{figure*}
\centering
\includegraphics[scale=0.7]{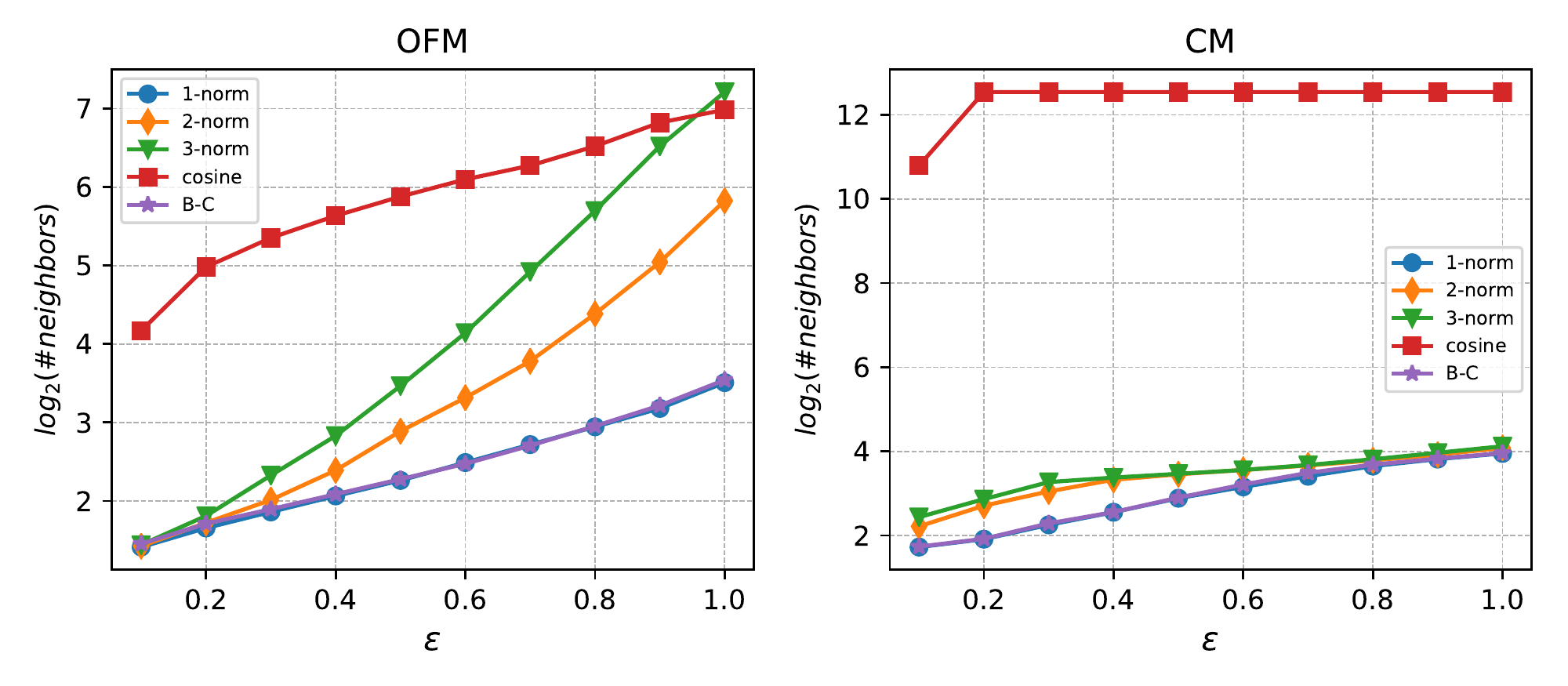}
\caption{Average number of neighbors of crystals determined by different values of $\varepsilon$ and similarity measures.}
\label{fig:nNeighbors}
\end{figure*}

For each crystal represented by OFM and CM, we determine its neighboring regions according to each similarity measure and $\varepsilon$. Next, for each crystal, we count the number of neighbors in its neighboring region. We take an average of the number of neighbors of all crystals in the dataset with a specific similarity measure and $\varepsilon$, as shown in Figure~\ref{fig:nNeighbors}. This figure shows that the 1-norm distance and B-C dissimilarity determine fewer neighbors than other measures with both descriptors. Therefore, these measures are more appropriate for approximating the energy function. In fact, the experimental results also show the improvement in prediction by using the 1-norm distance and B-C dissimilarity. Indeed, a small number of neighbors in a fixed radius determined by the 1-norm distance and B-C dissimilarity also indicates how these measures preserve the materials' distinctiveness.

\subsubsection{Similarity measure selection based on descriptors' characteristics}
\label{sec:desciptor_identity}
By collating the experimental results shown in Figure~\ref{fig:knn-acc-ofm} and Figure~\ref{fig:knn-acc-cm}, we see that there is significant improvement when using the 1-norm distance and B-C dissimilarity for OFM. Meanwhile, the improvement is not significant for CM, and these measures perform even worse than other measures in sample CM-G1. In other words, choosing appropriate similarity measures needs to be carried out strictly for OFM because this strongly affects the prediction performance. Meanwhile, for CM, inappropriately selecting similarity measures is tolerable. The underlying reason for this pertains to how distinct the instances are when they are represented by descriptors. The distinctiveness of data instances is indicated via (i) the number of dimensions in the representation, and (ii) whether the information encoded in representations can distinguish instances, which can be measured by estimating the variance of data.

Regarding (i), as shown in Table~\ref{tab:feature_var}, there are more dimensions in the OFM-based representation (1024 dimensions) than in the CM-based representation. Obviously, increasing the number of dimensions means enhancing the instances' distinctiveness. %Therefore, OFM is more appropriate than CM for representing materials with an emphasis on the materials' distinctiveness.
Therefore, the emphasis on the materials' distinctiveness in the OFM is stronger than it is in the CM. This results in the need to strictly select appropriate similarity measures for the OFM.

Regarding (ii), we examine the variance of data with different representations. Let $X_1, X_2,...,X_d$ be random variables that correspond to $d$ features in representations. The data variance is computed as $variance = \frac{1}{d}\sum_{i=1}^dVar(X_i)$, which is shown in Table~\ref{tab:feature_var}. The variance of data when represented by OFM is higher than when represented by CM. This also shows that materials represented by the OFM are more distinct than those represented by the CM, and that the OFM is more appropriate for representing materials in this case.

\begin{table}
\centering
\caption{Estimation of data variance where the data is represented by OFM and CM descriptors}
\label{tab:feature_var}
\begin{tabular}{c|c|c}\hline
Descriptor & \# dimensions & data variance \\ \hline
OFM & 1056 & 0.013\\
CM & 68 &  0.004\\\hline
\end{tabular}
\end{table}

Both the complexity of the energy function and the selection of similarity measures depend on the characteristics of the descriptors, as presented in Figure~\ref{fig:methodology}. If the representation essentially indicates the distinctiveness among instances, similarity measures that minimize the loss of instances' distinctiveness are more suitable for fitting a complex function of instances.

\subsection{Ridge regression}\label{sec:ridge}
Ridge regression is a parametric model that approximates the energy function by a linear function. In this method, the linear coefficients $\beta$ are estimated to minimize the penalized residual sum of squares, as follows:
%Ridge regression is a well-known linear regression method in which the linear coefficients $\beta$ are estimated to minimize the penalized residual sum of squares as follows:
\begin{equation}
RSS(\lambda) = (\textbf{y} - \textbf{X}\beta)^T(\textbf{y} - \textbf{X}\beta) + \lambda\beta^T\beta,
\end{equation}
where the matrix $\textbf{X}$ is the input data, and $\lambda \geq 0$ is a predefined parameter indicating an amount of coefficient shrinkage towards zero (weight decay). Ridge regression has the following closed-form solution:
\begin{equation}\label{eq:ridge_beta}
\hat{\beta} = (\textbf{X}^T\textbf{X} + \lambda\textbf{I})^{-1}\textbf{X}^T\textbf{y},
\end{equation}
where $\textbf{I}$ is the identity matrix.

This differs from locally approximating models such as the KNN model, insofar as parametric models are more generalized; they do not require all instances in the dataset to fit the energy function. Because of this generalization, it is possible to explore the nature of data from experimental results when fitting the model for the whole dataset. This incurs less risk of overfitting than the KNN interpretation.

We compare the performance (via RMSE, MAE, and the $R^2$) of ridge regression with the OFM and CM descriptors when predicting crystal formation energies. The most likely hyperparameter $\lambda$ is chosen by doing a grid search. We use ten-fold cross-validation to find the optimal $\lambda$. The optimal values of $\lambda$ for OFM and CM are 0.01 and 1.0, respectively. The prediction accuracies are presented in Table~\ref{tab:ridge_performance}. From the table, we see that by using ridge regression, the OFM descriptor outperforms the CM descriptor.

\begin{table}
\centering
\caption{Crystal formation energy prediction performance obtained using ridge regression for the OFM and CM descriptors}
\label{tab:ridge_performance}
\begin{tabular}{c|c|c}\hline
 Metric & OFM & CM \\ \hline
RMSE & \textbf{0.239}$\pm$\textbf{0.002} &  0.914$\pm$0.01\\
MAE & \textbf{0.184}$\pm$\textbf{0.002}& 0.722$\pm$0.006\\
$R^2$ & \textbf{0.97}$\pm$\textbf{0.01} & 0.556$\pm$0.01\\ \hline
\end{tabular}
%\vspace{1mm}
%\raggedright{\footnotesize{\textit{Note:} The best values of $\lambda$ for OFM and CM are 0.01 and 1.0, respectively.}}
\end{table}

%\subsection{Ridge and kernel ridge regression}\label{sec:ridge_krr}

%Ridge and kernel ridge regression (KRR) are parametric models that approximate the energy surface by closed-form expressions. They differ from locally approximating models such as KNN insofar as parametric models are more generalized; they do not take all instances in the dataset to fit the energy function. Because of this generalization, it is possible to explore the nature of data from experimental results when fitting ridge regression and KRR for the whole dataset. This incurs less risk of overfitting than the interpretation of KNN. Analogically, by interpreting experimental results, we can clarify the dependence on the descriptor and learning method when choosing appropriate kernel functions that are constructed from similarity measures.

\subsection{Kernel ridge regression}
\label{sec:krr}
KRR is the dual form of the ridge regression solution (see Appendix B). KRR aims to improve the performance of linear methods by mapping instances from the original space (Hilbert space) to a higher-dimensional space to acquire linearly separable patterns. Let $\phi$ be the mapping function. The pairwise dot product of instances in the new space is approximated by kernel functions $K(x_i, x_j) \approx \langle \phi(x_i), \phi(x_j) \rangle$, which form kernel matrices $\textbf{K}$ (i.e., Gram matrices). The radial basis function (RBF) kernel and Laplacian kernel, which are constructed from the 2-norm and 1-norm distances, respectively, have been widely used. The formulas for these kernels are as follows:

\begin{itemize}
\item RBF: $K(x_i, x_j) = exp(-\gamma||x_i - x_j||^2_2)$, where $||x_i-x_j||_2$ is the 2-norm distance between $x_i, x_j$, and $\gamma$ is a predefined scalar.
\item Laplacian: $K(x_i, x_j) = exp(-\gamma||x_i-x_j||_1)$, where $||x_i-x_j||_1$ is the 1-norm distance between $x_i, x_j$.
\end{itemize}
In addition, we can modify existing kernels by replacing the 1-norm and 2-norm distances by other similarity measures., we considered several derivations of RBF and Laplacian kernels using the 3-norm, cosine distances and B-C dissimilarity as follows:
\begin{itemize}
\item $Kernel_{3-norm}$: $K(x_i, x_j) = exp(-\gamma||x_i - x_j||_3)$ where $||x_i - x_j||_3$ is the 3-norm distance between $x_i$ and $x_j$.
\item $Kernel_{cosine}$: $K(x_i, x_j) = exp(-\gamma \times d_{cosine}(x_i,x_j))$ where $d_{cosine}$ is the cosine distance between $x_i$ and $x_j$
\item $Kernel_{B-C}$: $K(x_i, x_j) = exp(- \gamma \times d_{B-C}(x_i, x_j))$ where $d_{B-C}$ is the Bray-Curtis dissimilarity between $x_i$ and $x_j$.
\end{itemize}

We compare the crystal formation energy prediction performance by using KRR with different kernel functions (as listed above) and descriptors. The results are given in Table~\ref{tab:krr_performance}. In KRR, the most likely hyperparmeters $\lambda$ and $\gamma$ are selected based on a grid search with ten-fold cross-validation. The results are given in Table~\ref{tab:krr_hyperparameters}. Table~\ref{tab:krr_performance} shows that the Laplacian kernel and $Kernel_{B-C}$ outperforms the others with both OFM and CM descriptors.
 
\begin{table}
\centering
\caption{Selection of $\lambda$ and $\gamma$ for the model in KRR based on a grid search}
\label{tab:krr_hyperparameters}
\begin{tabular}{c|l|c|c} \hline
Descriptor & Kernel & $\lambda$ & $\gamma$ \\ \hline
\multirow{5}{*}{OFM} & RBF & $10^{-4}$ & $10^{-2}$\\ 
& Laplacian & $10^{-3}$ & $10^{-2}$\\
& $Kernel_{3-norm}$ & $1.0$ & $1.0$ \\
& $Kernel_{cosine}$ & $10^{-5}$ & $10^{-1}$\\
& $Kernel_{B-C}$ & $10^{-4}$ & $10^{-1}$\\ \hline
\multirow{5}{*}{CM} & RBF & $10^{-3}$ & $10.0$\\
& Laplacian & $10^{-2}$ & $1.0$\\
& $Kernel_{3-norm}$ & $10^{-1}$ & $10.0$\\
& $Kernel_{cosine}$ & $10^{-5}$ & $10^{-2}$\\
& $Kernel_{B-C}$ & $10^{-2}$ & $10^{2}$\\ \hline
\end{tabular}
\end{table}

%\begin{table}
%\centering
%\caption{Crystal formation energy prediction performance by KRR with different kernel functions and descriptors}
%\label{tab:krr_performance}
%\begin{tabular}{c|c|c|c|c|c|c}\hline
%& Metric & RBF & Laplacian & $Kernel_{3-norm}$ & $Kernel_{cosine}$ &  $Kernel_{B-C}$\\ \hline
%\multirow{3}{*}{OFM}& RMSE & 0.158$\pm$0.002  &  \textbf{0.108$\pm$0.002}  & 0.429$\pm$0.006  & 0.729$\pm$0.19  & \textbf{0.109$\pm$0.003}\\
%& MAE & 0.113$\pm$0.001  &  \textbf{0.067$\pm$0.001}  & 0.323$\pm$0.004   & 0.289$\pm$0.046 & \textbf{0.067$\pm$0.001}\\
%& $R^2$ & 0.987$\pm$ 0.001 &  \textbf{0.994$\pm$0.001}   & 0.902$\pm$0.003  & 0.521$\pm$0.272 & \textbf{0.994$\pm$0.001}\\ \hline
%\multirow{3}{*}{CM} & RMSE & 0.394$\pm$0.018 &  \textbf{0.319$\pm$0.008}  &  0.395$\pm$0.013  & 1.071$\pm$0.138 & \textbf{0.328 $\pm$0.011}\\
%& MAE &  0.245$\pm$0.006  &  \textbf{0.194$\pm$0.003}  & 0.246$\pm$0.005  & 0.671$\pm$0.01 & \textbf{0.19$\pm$0.004}\\
%& $R^2$ & 0.916$\pm$0.008 & \textbf{0.946$\pm$0.003}  &  0.917$\pm$0.005  & 0.285$\pm$0.208 &  \textbf{0.942$\pm$0.004}\\ \hline
%\end{tabular}
%\end{table}

\begin{table}
\centering
\caption{Crystal formation energy prediction performance by KRR with different kernel functions and descriptors}
\label{tab:krr_performance}
\begin{tabular}{c|l|c|c|c} \hline
Descriptor & Kernel & RMSE & MAE & $R^2$ \\ \hline
\multirow{5}{*}{OFM} & RBF & 0.158$\pm$0.002 & 0.113$\pm$0.001 & 0.987$\pm$0.001\\
& Laplacian & \textbf{0.108$\pm$0.002} & \textbf{0.067$\pm$0.001}  & \textbf{0.994$\pm$0.001}\\
& $Kernel_{3-norm}$ & 0.429$\pm$0.006 & 0.323$\pm$0.004 & 0.902$\pm$0.003 \\
& $Kernel_{cosine}$  & 0.729$\pm$0.19 & 0.289$\pm$0.046 & 0.521$\pm$0.272 \\
& $Kernel_{B-C}$ & \textbf{0.109$\pm$0.003} & \textbf{0.067$\pm$0.001} & \textbf{0.994$\pm$0.001} \\ \hline
\multirow{5}{*}{CM} & RBF  & 0.394$\pm$0.018 & 0.245$\pm$0.006 & 0.916$\pm$0.008 \\
& Laplacian & \textbf{0.319$\pm$0.008} & \textbf{0.194$\pm$0.003} & \textbf{0.946$\pm$0.003} \\
& $Kernel_{3-norm}$ & 0.395$\pm$0.013 & 0.246$\pm$0.005 & 0.917$\pm$0.005  \\
& $Kernel_{cosine}$ & 1.071$\pm$0.138 & 0.671$\pm$0.01 & 0.285$\pm$0.208 \\
& $Kernel_{B-C}$ & \textbf{0.328$\pm$0.011}& \textbf{0.19$\pm$0.01} & \textbf{0.942$\pm$0.004}  \\ \hline
\end{tabular}
\end{table}

\subsection{Model complexity}
The complexity of the model can be quantitatively interpreted by the degrees of freedom. The degrees of freedom are denoted by $df$ and defined as the number of freely varying parameters in the model (or function). In terms of model complexity, the greater the number of free parameters, the more complex the model is. For computation, the degrees of freedom are defined as the trace of the first derivatives of $\hat{\textbf{y}}$ according to $\textbf{y}$ as follows: 
\begin{equation}
df = tr \Bigg(\frac{\partial \hat{\textbf{y}}}{\partial \textbf{y}} \Bigg),
\end{equation}
where $\textbf{y}$ and $\hat{\textbf{y}}$ are the real target value and the estimated target value, respectively~\cite{kramer2007kernelizing}.

In ridge regression, because $\hat{\beta}$ is estimated with Equation~\ref{eq:ridge_beta}, we have
\begin{equation}
\hat{\textbf{y}} = \textbf{X}\hat{\beta} = \textbf{X}(\textbf{X}^T\textbf{X} + \lambda\textbf{I})^{-1}\textbf{X}^T\textbf{y}.
\end{equation}
Hence, the model’s degrees of freedom with a predefined $\lambda$, denoted by $df(\lambda)$, are estimated as $tr\Big(\textbf{X}(\textbf{X}^T\textbf{X} + \lambda\textbf{I})^{-1}\textbf{X}^T\Big)$.

In KRR, because $\hat{\beta} = \textbf{X}^T(\textbf{K} + \lambda\textbf{I})^{-1}\textbf{y}$ (see Equation~\ref{eq:a7} in Appendix B), we obtain
\begin{equation}
\begin{aligned}
\hat{\textbf{y}} = \textbf{X}\hat{\beta} &= \textbf{X}\textbf{X}^T(\textbf{K} + \lambda\textbf{I})^{-1}\textbf{y},\\
&= \textbf{K}(\textbf{K}+\lambda\textbf{I})^{-1}\textbf{y}.
\end{aligned}
\end{equation}
Therefore, the model’s degrees of freedom in KRR $df(\lambda)$ are estimated as $tr\Big( \textbf{K}(\textbf{K}+\lambda\textbf{I})^{-1} \Big)$.

The model’s degrees of freedom in ridge regression depend on the descriptor and the hyperparameter $\lambda$. Meanwhile, in KRR, this depends on the descriptor, similarity measure (used in kernel functions), and the hyperparmeters $\lambda$ and $\gamma$. Utilizing the most likely hyperparameters $\lambda$ and $\gamma$, as presented in Section~\ref{sec:ridge} and Table~\ref{tab:krr_hyperparameters}, we can estimate the model’s degrees of freedom in ridge and kernel ridge regression corresponding to each descriptor and kernel function. This is shown in Table~\ref{tab:dof}.

\begin{table}
\centering
\caption{Estimating the model’s degrees of freedom in ridge regression and KRR with different descriptors and kernels}
\label{tab:dof}
\begin{tabular}{l|l|c}\hline
Method & Descriptor and kernel & $df(\lambda)$ \\ \hline
\multirow{2}{*}{Ridge} & OFM & 441.87\\
& CM & 34.97\\ \hline
\multirow{5}{*}{KRR}& OFM + RBF & 2132.02\\
& OFM + Laplacian & \textbf{4087.08}\\
& OFM + $Kernel_{3-norm}$ &  1790.23\\
& OFM + $Kernel_{cosine}$ & 1981.61\\
& OFM + $Kernel_{B-C}$ & \textbf{4843.64}\\ \hline
\multirow{5}{*}{KRR} & CM + RBF & 711.06\\
& CM + Laplacian & \textbf{3018.84}\\
& CM + $Kernel_{3-norm}$ & 2819.88\\
& CM + $Kernel_{cosine}$ & 47.85\\
& CM + $Kernel_{B-C}$ & \textbf{3957.95}\\ \hline
\end{tabular}
\end{table}

\subsection{Descriptor selection based on model complexity}
As discussed in Sections~\ref{sec:visualization} and~\ref{sec:optimalk}, the energy function of crystals is complex. Hence, to appropriately fit the energy function, we should choose models with high complexity~\cite{exterkate2011modelling}. This means we need to select models that have high degrees of freedom.

Relying on the degrees of freedom of ridge regression, we can evaluate the appropriateness of the OFM and CM descriptors for approximating the energy function. From Table~\ref{tab:dof}, we see that the model’s degrees of freedom in ridge regression when using OFM ($df(\lambda) = 441.87$) are greater than when using CM($df(\lambda) =34.97$). Therefore, the linear model approximated from the OFM representation of materials has a higher complexity than that approximated from the CM representation. This may explain why representing materials by OFM results in better performance when predicting the formation energies of materials compared to representing materials by CM, as shown in Table~\ref{tab:ridge_performance}.

\subsection{Kernel selection  based on model complexity}\label{sec:kernel_select}
By estimating the degrees of freedom of KRR, we can select not only the appropriate descriptor but also the appropriate kernel. In KRR, the use of a Laplacian and $Kernel_{B-C}$ results in higher complexity of the model than using other kernels. This explains why a Laplacian and $Kernel_{B-C}$ perform better than others with both OFM and CM in terms of predicting crystal formation energies, as shown in Table~\ref{tab:krr_performance}. As mentioned in Section~\ref{sec:krr}, the function $\phi$ maps instances in the data to a higher-dimensional space, which enhances the distinctiveness among materials. In fact, kernel functions can be treated as generalized similarity measures in the new space of these instances~\cite{pekalska2001generalized}. Therefore, supposing that if we fit the energy function based on instances in the higher-dimensional space by KNN, kernel functions, which minimize the loss of materials' distinctiveness, will result in better performance, as discussed in Section~\ref{sec:knn}. The Laplacian and $Kernel_{B-C}$, which are monotonic functions of the 1-norm distance and B-C dissimilarity, also minimize the loss of materials' distinctiveness. Thus, these kernels are more appropriate than the others for fitting the energy function in the new space. In fact, KRR approximates this function in the same manner as KNN, but the former uses the model with a closed-form expression rather than locally fitting models, and we see the advantages of using the Laplacian and $Kernel_{B-C}$.

Relying on the model complexity estimation presented in Table~\ref{tab:dof} when using the Laplacian and $Kernel_{B-C}$, the model using OFM has higher complexity than the model using CM. This induces better performance with the KRR compared to using OFM.

Through the interpretations presented above, we clarify the association between the emphasis on materials' distinctiveness (which includes reflecting the instances' distinctiveness of the descriptors and minimizing the loss of this distinctiveness of similarity measures) and the model complexity. This forms the basis for selecting appropriate descriptors and kernel functions (or similarity measures) to effectively mine material data.

\section{Conclusion}
This paper introduces a protocol for interpreting experimental results when mining material data in a case study that predicts crystal formation energies by KNN, ridge regression, and KRR. Through empirical and theoretical interpretations of the prediction performance from multiple perspectives, we found the dependence among descriptors, similarity measures, and learning methods. This forms the basis for model selection to effectively mine materials data. In case these factors simultaneously reflect the nature of data, high performance can be obtained with mining tasks. Through the case study, we found that descriptors that suitably reflect the materials' distinctiveness and similarity measures that minimize the loss of this distinctiveness result in better performance when predicting the formation energies of materials. This research can serve as the groundwork for future studies regarding the use of machine learning methods for mining material data.

\section*{Acknowledgements}
This work was partly supported by PRESTO and the ``Materials Research by Information Integration'' initiative (MI$^2$I) project of the Support Program for Starting Up Innovation Hub, by the Japan Science and Technology Agency (JST) and the Elements Strategy Initiative Project under the auspices of MEXT, and also by MEXT as a social and scientific priority issue (Creation of New Functional Devices and High-Performance Materials to Support Next-Generation Industries; CDMSI) to be tackled using a post-K computer.

%An unnumbered section, e.g.\ \verb"\section*{Acknowledgements}", may be used for thanks, etc.\ if required and included \emph{in the non-anonymous version} before any Notes or References.

\section*{Disclosure statement}

The authors declare that they have no conflict of interest.

\bibliographystyle{tfnlm}
\bibliography{references}

\section{Appendices}
\appendix
\section{Example of approximating complex functions using KNN regression}
\begin{figure}
\centering
\includegraphics[scale=0.6]{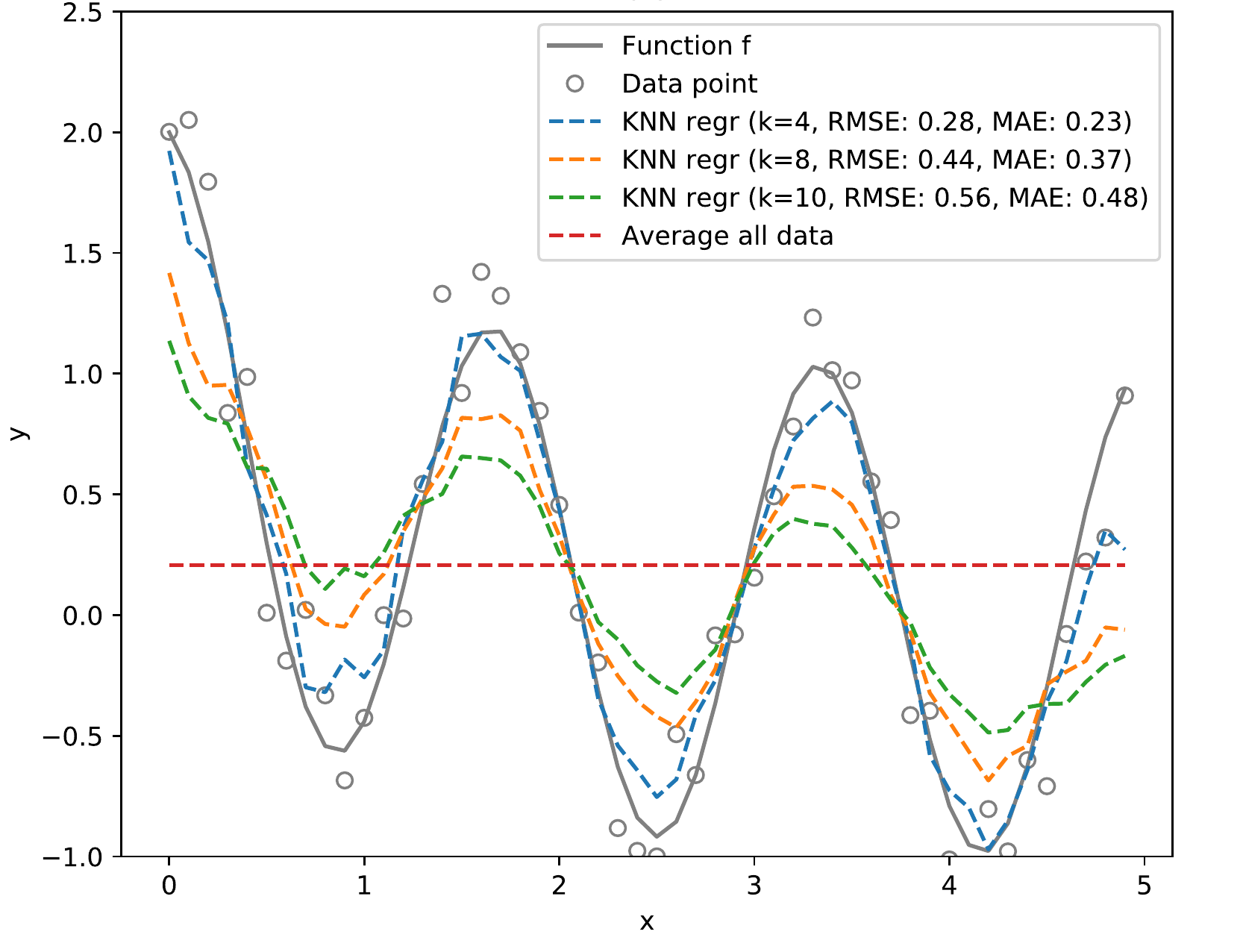}
\caption{Example of fitting a flexible function (with many extreme points and high curvature) over one-dimensional data points using KNN regression with various values of $k$ ($k=4,8,10$). The performance of the fitting is evaluated via RMSE and MAE.}
\label{fig:fitExample}
\end{figure}

KNN regression approximates a global function of all data points $f(x)$ by averaging the neighbors of each point. To illustrate how KNN approximates a flexible and complex function, we consider a univariate dataset, in which the distance between data points is estimated by $|x_i - x_j|$, where $x_i$ and $x_j$ are scalar. Suppose that we generate a sample of data points from the following function:

\begin{equation}
\begin{aligned}
f(x) & = e^{-x} + \cos(1.2 \times \pi x) + \epsilon, \\
\epsilon &\sim \mathcal{N}(\mu, \sigma^2),
\end{aligned}
\end{equation}
where $x$ indicates data points that are scalar, and $\epsilon$ is the noise generated from a normal distribution $\mathcal{N}(\mu, \sigma^2)$. For each data point $x$, we use KNN to approximate its target value, where the number of neighbors ($k$) is 4, 8, and 10. Setting $k$ by even numbers ensures that the number of neighbors on both sides of each data point are equal. The accuracy at different values of $k$ is evaluated via RMSE and MAE, and the results are given in Figure~\ref{fig:fitExample}.

Two remarks are worth noting regarding the approximation of a complex function by KNN: (i) it is difficult to estimate the function values precisely at data points that are extreme points of the functions; (ii) the use of a smaller value of $k$ ($k = 4$) gives a better approximation than the use of a larger value, and a small $k$ is particularly suitable for approximating extreme points. 

\section{Kernel ridge regression -- dual form of ridge regression}

We rewrite the optimization problem for ridge regression as

\begin{equation}
\begin{aligned}
\underset{\beta, \textbf{r}}{minimize} &\quad \frac{1}{2}\Bigg(||\textbf{r}||^2_2 + \lambda||\beta||^2_2 \Bigg)\\
\text{subject to} &\quad \textbf{r} = \textbf{X}\beta - \textbf{y}
\end{aligned}
\end{equation}

The solution is equivalent to

\begin{equation}
\begin{aligned}
&\min_{\beta, \textbf{r}}\max_{\alpha}L(\beta, \textbf{r}, \alpha)\\
=& \min_{\beta, \textbf{r}}\max_{\alpha}\Bigg(\frac{1}{2}||\textbf{r}||^2_2 + \frac{\lambda}{2}||\beta||^2_2 + \alpha^T(\textbf{r}-\textbf{X}\beta+y) \Bigg),
\end{aligned}
\end{equation}

where $L(\beta, \textbf{r}, \alpha)$ is the Lagrangian function. We solve the minimization problem by setting to zero the first derivatives of the Lagrangian function according to $\beta$ and $\textbf{r}$:

\begin{equation}\label{eq:a3}
\begin{aligned}
\frac{\partial L}{\partial \beta}(\beta, \textbf{r}, \alpha) &= 0 \Rightarrow \lambda\beta - \textbf{X}^T\alpha = 0 \Rightarrow \hat{\beta} = \frac{1}{\lambda} \textbf{X}^T \alpha \\
\frac{\partial L}{\partial \textbf{r}}(\beta, \textbf{r}, \alpha) &= 0 \Rightarrow \textbf{r} + \alpha=0 \Rightarrow \hat{\textbf{r}}=-\alpha
\end{aligned}
\end{equation}

Plugging $\hat{\beta}$ and $\hat{\textbf{r}}$ into the Lagrangian function obtains

\begin{equation}
\begin{aligned}
L(\hat{\beta}, \hat{\textbf{r}}, \alpha) &= \frac{1}{2}||\alpha||^2_2 + \frac{1}{2\lambda}||\textbf{X}^T\alpha||^2_2 + \alpha^T(-\alpha-\frac{1}{\lambda}\textbf{X}\textbf{X}^T\alpha + \textbf{y}) \\
&= -\frac{1}{2}||\alpha||^2_2 - \frac{1}{2\lambda}\alpha^T\textbf{X}\textbf{X}^T\alpha + \alpha^T\textbf{y}
\end{aligned}
\end{equation}

Now, the dual problem is $\max_{\alpha}L(\hat{\beta}, \hat{\textbf{r}}, \alpha)$, which is equivalent to the following (noting that $\lambda \geq 0$):

\begin{equation}\label{eq:a4}
\min_{\alpha}\Bigg(\frac{1}{2}\alpha^T(\textbf{K} + \lambda\textbf{I})\alpha - \lambda\alpha^T\textbf{y} \Bigg),
\end{equation}
where $\textbf{K} = \textbf{X}\textbf{X}^T$ is called the kernel matrix. To obtain $\alpha$, we also set the first derivatives of the dual objective function to zero, to obtain

\begin{equation}
\begin{aligned}
& (\textbf{K} + \lambda\textbf{I})\alpha - \lambda\textbf{y} = 0\\
\Rightarrow& \alpha = \lambda(\textbf{K} + \lambda\textbf{I})^{-1}\textbf{y}
\end{aligned}
\end{equation}
Based on Equation~\ref{eq:a3}, we obtain

\begin{equation}\label{eq:a7}
\beta = \textbf{X}^T(\textbf{K} + \lambda\textbf{I})^{-1}\textbf{y}
\end{equation}

\end{document}